\documentclass[10pt,journal,compsoc]{IEEEtran}
\usepackage{amsmath,amssymb,amsfonts}
\usepackage{algorithmic}
\usepackage{graphicx}
\usepackage{textcomp}
\usepackage[table,xcdraw]{xcolor}
\usepackage{array}
\usepackage{multirow}
\usepackage{enumerate}
\usepackage{url}
\usepackage{multirow}
\usepackage{booktabs}
\usepackage{xcolor}

\usepackage[normalem]{ulem}
\useunder{\uline}{\ul}{}

\ifCLASSOPTIONcompsoc
  \usepackage[nocompress]{cite}
\else
  \usepackage{cite}
\fi
\ifCLASSINFOpdf
\else
\fi
\begin{document}

%
\title{Critically examining the Domain Generalizability \\ of Facial Expression Recognition models}
%
%
%
%

\author{Varsha~Suresh,
        Gerard~Yeo,
        and~Desmond~C.~Ong,~\IEEEmembership{Member,~IEEE~Computer~Society}
\IEEEcompsocitemizethanks{
\IEEEcompsocthanksitem V. Suresh is with the Department of Computer Science, National University of Singapore
\IEEEcompsocthanksitem G. Yeo is with the Institute of Data Science, Integrative Sciences and Engineering Programme, National University of Singapore
\IEEEcompsocthanksitem D. C. Ong is with the Department of Psychology at The University of Texas at Austin, and was previously with the Department of Information Systems and Analytics at the National University of Singapore \protect\\ 
E-mail: desmond.ong@utexas.edu}%
}

%
%

\markboth{IEEE TRANSACTIONS ON AFFECTIVE COMPUTING, MANUSCRIPT ID}%
{Suresh \MakeLowercase{\textit{et al.}}: Systematic Review of Domain Generalization in FER}
%



\IEEEtitleabstractindextext{%
\begin{abstract}
    Facial Expression Recognition is a commercially-important application, but one under-appreciated limitation is that such applications require making predictions on out-of-sample distributions, where target images have different properties from the images the model was trained on. How well---or how badly---do facial expression recognition models do on unseen target domains? We provide a systematic and critical evaluation of transfer learning---specifically, domain generalization---in facial expression recognition. Using a state-of-the-art model with twelve datasets (six collected in-lab and six ``in-the-wild"), we conduct extensive round-robin-style experiments to evaluate classification accuracies when given new data from an unseen dataset. We also perform multi-source experiments to examine a model's ability to generalize from multiple source datasets, including (i) within-setting (e.g., lab to lab), (ii) cross-setting (e.g., in-the-wild to lab), and (iii) leave-one-out settings. Finally, we compare our results with three commercially-available software. 
    We find sobering results: the accuracy of single- and multi-source domain generalization is only modest. Even for the best-performing multi-source settings, we observe average classification accuracies of 65.6\% (range: 34.6\%-88.6\%; chance: 14.3\%), corresponding to an average drop of 10.8 percentage points from the within-corpus classification performance (mean: 76.4\%). 
    We discuss the need for regular, systematic investigations into the generalizability of affective computing models and applications. 
    %
\end{abstract}

\begin{IEEEkeywords}
Affective Computing; Facial Expressions; Ethical/Societal Implications; Transfer Learning; Domain Generalization
\end{IEEEkeywords}}

\maketitle

\IEEEdisplaynontitleabstractindextext

\IEEEpeerreviewmaketitle


\section{ Introduction }

Recent technological advancements in machine learning and computer vision have resulted in the proliferation of facial expression recognition technologies that claims to accurately recognize emotions from facial expressions. 
Startups and large technology firms alike offer technologies that claim to predict what a person is feeling, based on their facial expressions alone---often, from a still image with no context. And more and more companies are using these services to power their own ``emotionally-intelligent" offerings.

Intuitively, such a technological capability seems plausible and well within the capabilities of modern deep learning methods. Consider object recognition: deep learning models seem to be able to classify scores of objects at accuracies and speeds that outperform humans \cite{russakovsky2015imagenet}, and at mind-boggling scales. Moreover, people are excellent at reading emotions from faces, and this ease might confer a confidence that machines could learn to do this task as well. 

Yet, there has been growing evidence that facial expression recognition is far from being a solved problem, and for many reasons. A recent comprehensive review argues that the inference of emotional states from facial expressions may itself be problematic, because there is considerable variability: there exists not a one-to-one mapping between emotions and ``facial configurations", but a complex, many-to-many mapping \cite{barrett2019emotional}.
Directly citing the previous paper, a report from the AI Now Institute even called for regulators to ban the use of affect recognition in decisions that could impact people's lives, citing the ``contested scientific foundations of affect recognition technology" \cite{crawford2019ai}. 

Given the increasing ubiquity of facial expression recognition technology, it is essential to critically examine the scientific validity of such technology. Here, we conduct a thorough and systematic review of one aspect of validity: how well do these models perform when asked to make predictions on novel data?

An under-appreciated limitation of these models is external generalizability. These state-of-the-art systems, especially deep learning-based systems \cite{li2020deep}, are trained using supervised learning, which involves minimizing the classification error on a dataset of faces and emotion labels. Even though it is standard practice to ``hold out" a portion of the dataset to evaluate the test accuracy of a classifier, such cross-validation techniques provide a measure of the expected classification accuracy only for new samples \emph{drawn from the same distribution as the training data}. But in many applications, the data that one wishes to make predictions on may be very different---along both emotion-relevant as well as emotion-irrelevant dimensions---from the training distribution. Multiple factors such as differences in the features that are predictive of emotions when the expressions are posed or generated spontaneously (an important distinction for facial expression recognition datasets),  
changes in facial feature distributions due to ethnicity, gender, and age; or even ``emotion-irrelevant" changes in pose, background complexity, and illumination, can all contribute to what are called ``dataset shifts", ``feature-space shifts", or ``distribution shifts", which can impact a model's performance on different datasets \cite{zhang2019recent,koh2021wilds, rabanser2019failing,yu2021empirical,sagawa2019distributionally,wang2022generalizing}.



Using a model trained in one context (e.g., in one domain or on a specific task) to make predictions in another context (e.g., to a different domain or task) is broadly known as transfer learning \cite{pan2009survey, zhuang2020comprehensive, weiss2016survey}, where the knowledge in a trained model is ``transferred". Here, we focus on the particular case of transfer learning called \emph{domain generalization}\cite{zhou2022domain,wang2022generalizing}, where the facial expression recognition models are performing the same task (facial expression recognition), but in a new domain---a \emph{target} dataset that is different from the \emph{source} dataset that it was trained on. Specifically, in this investigation, we restrict the model to not see any training examples from the target dataset before having to make predictions on the target dataset (as opposed to domain adaptation, in which the model is given examples from the target dataset to learn to `adapt' to the target domain). We chose this bar because domain generalization is precisely what commercially available software claims to do: users can simply upload any image (of a face) without any context, and these software would make a prediction of the emotion ``present" in the image.



\begin{figure}[tbh!]
\centering
\includegraphics[width=1\columnwidth]{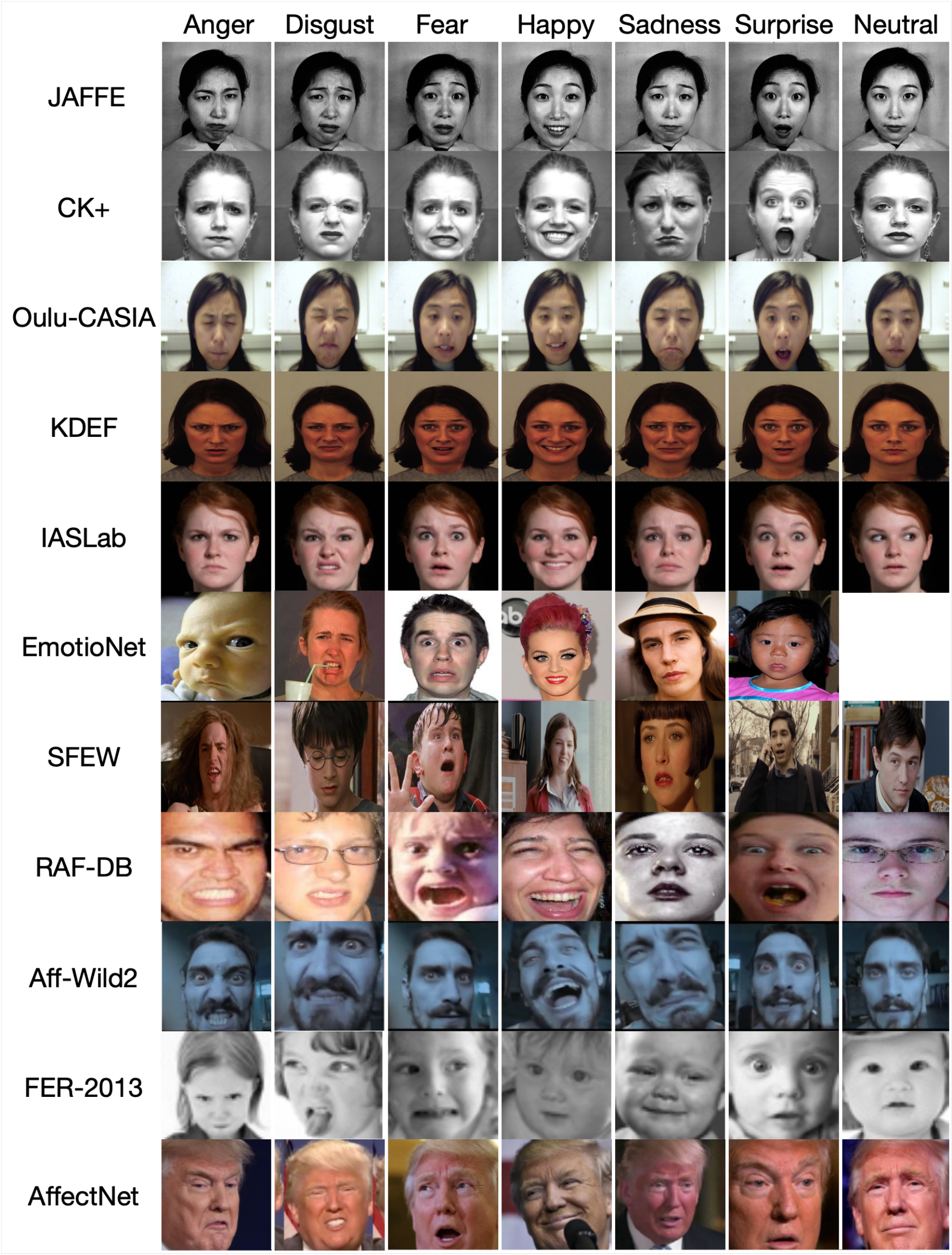}
\caption{Sample images from 11 of the 12 datasets (GEMEP does not allow image publication). The top 5 datasets are collected in the lab, and the bottom 6 are collected ``\emph{in-the-wild}". EmotioNet does not have neutral images.
In accordance with the various EULAs, the image ids from the KDEF dataset are: AF01AFS, AF01ANS, AF01DIS, AF01HAS, AF01NES, AF01SAS, AF01SUS; CK+, S55; Oulu-CASIA, P002; IASLab, F14. For JAFFE, figures are taken from original paper. 
}
\label{fig:exampleImages}
\end{figure}

In this paper, we conduct the largest-to-date systematic empirical evaluation of domain generalization using twelve facial expression recognition datasets, six of which are collected in the laboratory, and six are collected \emph{in-the-wild}, i.e., with more naturalistic variation (see Fig. \ref{fig:exampleImages}). 
We used three state-of-the-art deep learning-based models: The first, ResNet50 \cite{he2016deep} with pre-trained weights from VGGFace2 \cite{cao2018vggface2}; the second: Inception-ResNet \cite{szegedy2017inception} pre-trained on CASIA-WebFace dataset \cite{yi2014learning}; and ResNet-50 with entropy regularisation as a domain generalization technique \cite{zhao2020domain}. For brevity, we report only the results of the first model in the main text; the latter two models qualitatively replicate the results of the first model and demonstrate that our results are not specific to one model architecture; these are reported in full in the Appendix. 

We conducted three experiments to evaluate the generalization performances of facial expression recognition models. In Experiment 1, we investigate the cross-corpus domain generalization performance when the model is trained on one source dataset and evaluated on a different target dataset---with twelve datasets, this produces (12*11=) 132 permutations for which the source and target are different. 

In Experiment 2, we evaluate multiple-source training, which is a common practice for domain generalization approaches in machine learning fields like Natural Language Processing \cite{talmor2019multiqa,su2019generalizing} and Computer Vision \cite{wang2019cross,ji2019cross,dias2022cross}. This is to empirically verify if training on multiple facial expression recognition datasets improves model generalization. We perform experiments in a variety of setups: 1) within-setting (e.g., train on in-lab data, test on in-lab data), 2) cross-setting (e.g., train on in-lab data, test on ``in-the-wild'' data), and 3) a leave-one-out setup where we train on eleven datasets and evaluate on the last dataset. 

In Experiment 3, we evaluate the generalization of three widely available commercial Application Programming Interfaces, or APIs (Amazon Rekognition, Megvii Face++, and Microsoft Azure) on the datasets included in this study, and compare their results with our Experiments 1 and 2. 
We end with a frank discussion about the implications of our work.

\section{  Related Work  }


\subsection{ Psychological Evidence }

Humans are emotionally expressive, and the most visible expressions of emotions occurs on our faces \cite{elfenbein2002universality, russell2003facial, ekman1993facial}. Facial expressions provide important information about others' emotions: for example, a smile or a frown could lead one to infer that a person is feeling happy, or angry, respectively. For decades, psychologists have studied how people recognize emotions from observed facial configurations \cite{ekman1987universals, ekman1972universals, izard1971face}. Early work also seemed to support a theory that some emotions may be universally expressed and recognized across cultures \cite{ekman1971constants}, and this thinking has been very influential in computer science. 
However, other recent work, such as a meta-analysis conducted by \cite{elfenbein2002universality} found that although these ``basic emotions" (namely, anger, disgust, fear, joy, sadness, and surprise \cite{ekman1971constants}) were recognized at better-than-chance level, there were still cultural variations in the accuracy of emotion recognition, and these accuracies are far from being perfect (depending on the specific experiment). Another meta-analysis \cite{duran2021emotions} suggested that the correlations between the ``prototypical" facial configurations associated with emotions (as operationalized by Action Unit profiles) and actual emotional experiences is low, between $r$=13. to $r$=.30 depending on the analysis.
This issue of variability was examined in even greater detail by \cite{barrett2019emotional}, who critically examined the evidence for the scientific validity of inferring emotional states from facial expressions, arguing that facial configurations are not ``fingerprints" that signal particular emotional states across contexts and cultures: instead these associations are variable, and heavily dependent on context.

The recent psychological evidence suggests that emotion recognition from faces is not a straightforward task: faces are not expressed or interpreted the same way across contexts, or across individuals from different cultures. There is obviously information in facial expressions that allow people to accurately perceive the underlying emotions, but in real life, people often make these inferences with more information (information from other modalities, knowledge about the context, cultural knowledge). Thus, the machine learning challenge of inferring emotions from facial expressions alone should be thought of more as an underdetermined inference problem \cite{ong2015affective}, instead of being similar to object classification. We return to this point in the Discussion.

\subsection{Automated Facial Expression Recognition}

That said, 
automated facial expression recognition has been explored over several decades and using many approaches \cite{sariyanidi2015automatic, corneanu2016survey}. Early approaches to facial expression recognition used hand-crafted features: some examples are Active Appearance Model (AAM) \cite{cheon2009natural}, Scale Invariant Feature Transform (SIFT) \cite{luo2007person}, Local Binary Patterns (LBP) \cite{shan2009facial}, and Histogram of Oriented Gradients (HOG) \cite{chen2014facial} which is used, for example, by Affectiva's AffDex \cite{mcduff2016affdex}. These features are then subsequently used to classify an image into one of several emotion categories. 

In recent years, the availability of large facial expression recognition datasets and the increase in computational capabilities have enabled the training of end-to-end deep learning models such as CNN-based architectures like ResNet \cite{he2016deep} and AlexNet \cite{krizhevsky2012imagenet}. Deep learning approaches have emerged as the most successful approach to facial expression recognition \cite{li2020deep}. The modal approach today start with models pre-trained on large facial recognition datasets such as VGGFace2 \cite{cao2018vggface2}, and then fine-tune them to facial expression recognition using a facial expression recognition dataset \cite{wang2020region,behera2019cnn,zeng2022face2exp}.

\begin{table*}
\centering
\resizebox{\textwidth}{!}{
\begin{tabular}{c|c|c|c|c}
\textbf{Citation}  & \textbf{\# datasets} & \textbf{Single-Source}  & \textbf{Multi-Source} & \textbf{Transfer} \\ 
\hline
Gu \textit{et al.,} 2012 \cite{gu2012facial}                     & 2             & All source-target combinations       & -                              & None                               \\
Li \textit{et al.,} 2017 \cite{li2017reliable}                   & 2                    & All source-target combinations       & -                              & None                               \\
Yang \textit{et al.,} 2020 \cite{yang2020effects}                & 2                    & All source-target combinations       & -                              & None                               \\
Cruz\textit{ et al., }2014 \cite{cruz2014one}                    & 2                    & All source-target combinations       & -                              & None                               \\
Zhang\textit{ et al., }2015 \cite{zhang2015facial}               & 2                    & All source-target combinations       & -                              & None                               \\
Li \textit{et al.,} 2020 \cite{li2020facial}                     & 2                    & All source-target combinations       & -                              & None                               \\
Mavani \textit{et al.,} 2017 \cite{mavani2017facial}             & 2                    & One fixed source                     & -                              & None                               \\
Meng\textit{ et al.,} 2017 \cite{meng2017identity}               & 2                    & All source-target combinations       & -                              & None                               \\
Shan \textit{et al.,} 2009\cite{shan2009facial}                  & 3                    & One fixed source                     & -                              & None                               \\
Ali \textit{et al.,} \cite{ali2016boosted}                       & 3                    & All source-target combinations       & -                              & None                               \\
Zhu \textit{et al.,} 2015 \cite{zhu2015atransfer}                & 3                    & One fixed source                     & -                              & None                               \\
Hasani \textit{et al.,} 2017 \cite{hasani2017spatio}             & 3                    & -                                    & Leave-one-out                  & None                               \\
Mayer \textit{et al.,} 2014 \cite{mayer2014cross}                & 3                    & All source-target combinations       & -                              & None                               \\
Liu \textit{et al.,} 2015 \cite{liu2015inspired}                 & 3                    & Two datasets fixed as source         &                                & None                               \\
Lopes \textit{et al.,} 2017 \cite{lopes2017facial}               & 3                    & One fixed source                     & -                              & None                               \\
Wen \textit{et al.,} 2017 \cite{wen2017ensemble}                 & 4                    & One fixed source                     & -                              & None                               \\
da Silva \textit{et al.,} 2015 \cite{da2015effects}                 & 4                    & All source-target combinations       & -                              & None                               \\
Xie \textit{et al.,} 2019 \cite{xie2019sparse}                   & 4                    & All source-target combinations       & -                              & None                               \\
Barros \textit{et al.,} 2020 \cite{barros2020facechannel}        & 4                    & One fixed source                     & -                              & None                               \\
Zeng \textit{et al.,} 2018 \cite{zeng2018facial}                 & 7                    & -                                     & Two datasets fixed as source   & None                               \\
Mollahosseini \textit{et al.,} 2016\cite{mollahosseini2016going} & 7                    & -                                    & Leave-one-out                  & None                               \\
Zavarez \textit{et al.,} 2017 \cite{zavarez2017cross}            & 7                    & -                                    & Non-exhaustive leave-one-out   & None                               \\ 
\hline
Yan \textit{et al.,} 2016 \cite{yan2016transfer}                 & 3                    & All source-target combinations       & -                              & Adaptation                         \\
Miao \textit{et al.,} 2012 \cite{miao2012cross}                  & 3                    & One fixed source                     & Non-exhaustive leave-one-out   & Adaptation                         \\
Zhou \textit{et al.,} 2020 \cite{zhou2020uncertainty}            & 4                    & One fixed source                     & Only one combination           & Adaptation                         \\
Yan \textit{et al.,} 2019 \cite{yan2019cross}                    & 4                    & Non-exhaustive combinations          & -                              & Adaptation                         \\
Zhu \textit{et al.,} 2016 \cite{zhu2016discriminative}           & 4                    & One fixed source                     & -                              & Adaptation                         \\
Li \textit{et al.,} 2018 \cite{li2018deep}                       & 6                    & One fixed source                     & -                              & Adaptation                         \\
Li \textit{et al.,} 2021 \cite{li2021jdman}                      & 7                    & One fixed source                     & -                              & Adaptation                         \\
Chen \textit{et al.,} 2021 \cite{chen2021cross}                  & 7                    & All source-target combinations       & -                              & Adaptation                         \\
Li \textit{et al.,} 2020 \cite{li2020deeper}                     & 8                    & One fixed source                     & Only one combination           & Adaptation                         \\
                                 &   & (All combinations only for baseline) &                                &                                    \\ 
\hline
Wang \textit{et al.,} 2019 \cite{wang2019cross}                  & 4                    & -                                    & Non-exhaustive leave-one-out   & Generalization                     \\
Ji \textit{et al.,} 2019 \cite{ji2019cross}                      & 5                    & Non-exhaustive combinations          & Non-exhaustive leave-one-out   & Generalization                     \\
Dias \textit{et al.,} 2022 \cite{dias2022cross}                  & 8                    & -                                    & Three datasets fixed as source & Generalization                     \\ 
\hline \hline
\multirow{3}{*}{Current paper} & \multirow{3}{*}{12} & \multirow{3}{*}{All source-target combinations} & \multirow{3}{*}{Leave-one-out} & None \\
& & & & (Generalization \\
& & & & in Appendix)
\end{tabular}
}
\vspace{0.3cm}
\caption{Summary of previous research on cross-domain prediction in facial expression recognition, including the number of datasets they used, a brief description of their single-source and multi-source experiments, and whether they considered domain adaptation/generalization techniques (``None" indicates that the paper did not use any transfer learning techniques). Leave-one-out means that one dataset is used as target, while the remaining datasets are used as source datasets, and this is repeated for all target datasets.}
\label{tab:lit_review}
\end{table*}

\subsection{Cross-domain Facial Expression Recognition} \label{sec:cross_domain_fer}


Deep learning models are trained to minimize some objective function on a held-out subset of their labeled data, where the training data and the held-out data have similar distributions of features. Thus, these models will only work well in scenarios and applications where the new data for which predicted labels are desired are drawn from the same feature distribution as the data the models were trained on. 
It is important to understand how these models perform in, or ``transfer to", new contexts that have a different data distribution compared to the training data. 

\subsubsection{Domain Adaptation in Facial Expression Recognition} 
One such transfer task is formally known as domain adaptation. In domain adaptation, the aim is to improve the performance of a model that is trained using a \textbf{source} data from some domain (e.g., Dataset A) on an \textbf{target} data from a different domain (e.g., Dataset B), \emph{using some amount of data from the target domain}, which allows the model to learn to ``adapt" its learning to the target domain. 
A number of methods have been proposed  \cite{ganin2016domain,tzeng2017adversarial,xiao2021dynamic,fernando2013unsupervised}. 
Specifically for facial expression recognition, one class of approaches focus on re-weighting training instances based how different they are from the target samples \cite{miao2012cross, zhou2020uncertainty}. Other approaches focus on minimizing the discrepancy between the source and target distributions using methods such as subspace matching \cite{yan2016transfer}, or by incorporating discrepancy metrics such as Maximum Mean Discrepancy to the objective function \cite{li2018deep,li2020deeper,yan2019cross,zhu2016discriminative}. A third class of techniques use adversarial methods to ``un-learn" the domain-specific discriminative features \cite{chen2021cross,li2021jdman}. 
These approaches, however, assume that some amount of target data is available (either labeled or unlabeled), which allows the model to adapt to the target domain.


\subsubsection{Domain Generalization} 
A similar but stricter task tries to get the model to generalize to new domains where there is no target data available in any form. These are called \emph{domain generalization} approaches,
and have been studied widely outside of facial expression recognition  \cite{wang2022generalizing,zhou2022domain}.  
Some approaches focus on increasing the number and diversity of data samples either by data-augmentation \cite{volpi2018generalizing,yue2019domain,tobin2017domain} or synthetic data generation \cite{li2021progressive,anoosheh2018combogan,zhang2018mixup}. Another set of approaches uses data from multiple domains (i.e., multiple source datasets) to learn domain-invariant features using methods like adversarial losses \cite{shao2019multi,matsuura2020domain,li2018domain,zhao2020domain}, or moment-matching  \cite{zhang2021robust,mahajan2021domain}. 

Within facial expression recognition, there are only three papers that we know of that studied domain generalization techniques. 
\cite{ji2019cross} used dual feature-extractors to find domain-invariant features: one extractor minimized the distances between the features of samples across different domains that belong to the same class, while the second aimed to find distinguishing features amongst the classes. \cite{wang2019cross} used adversarial learning where a gradient-reversal layer was used to learn domain-invariant features. In addition, their model also minimized the feature distances between samples across multiple domains that belong to same class.  \cite{dias2022cross} used a triplet loss to minimise the distances between the feature vectors of samples from the same classes. 

Domain generalization is a difficult problem, but it needs to be solved to apply facial expression recognition in the real world. The main results in the current paper use a commonly-used deep-learning architecture for predicting across different domains; We specifically do not incorporate any generalization techniques (e.g., learning domain-invariant features) to simulate the performance of real-life applications that may not have considered generalization issues. But to also show how explicit domain-generalization training could improve our results, we adapted \cite{zhao2020domain}'s Entropy Regularization technique to facial expression recognition, and we report this model in the Appendix. 


\subsubsection{ Summary of cross-domain work }

We summarize the work on cross-domain facial expression recognition in Table \ref{tab:lit_review}. In addition to the domain adaptation and domain generalization papers cited above, we also include papers that report any cross-corpus prediction results and which did not use any specific transfer techniques \cite{gu2012facial,li2017reliable,yang2020effects,cruz2014one,zhang2015facial,li2020facial,mavani2017facial,meng2017identity,shan2009facial,ali2016boosted,zhu2015atransfer,hasani2017spatio,mayer2014cross,liu2015inspired,lopes2017facial,wen2017ensemble,da2015effects,xie2019sparse,barros2020facechannel,zeng2018facial,mollahosseini2016going,zavarez2017cross}. (These papers often report the results of their models on several target datasets as evidence for the generalizability of their methods. We note that these models were not trained or fine-tuned on the target dataset.)

In summarizing the large number of citations in Table \ref{tab:lit_review}, we wish to point out two key observations. First, the vast majority of the 34 papers we identified considered very few datasets, with only 9 papers \cite{zeng2018facial,mollahosseini2016going,zavarez2017cross,wang2019cross,ji2019cross,dias2022cross,li2021jdman,chen2021cross,li2020deeper}  that considered 5 or more datasets in their experiments, with the maximum number being eight datasets \cite{dias2022cross, li2020deeper}. But even amongst all these papers, there is not much systematicity in the source-target combinations that they considered, with many papers only examining a small subset of the possibilities. For example, of the papers that examined 5 or more datasets, only \cite{chen2021cross} considered all source-target combinations for single-source transfer, and only \cite{mollahosseini2016going} considered all leave-one-out combinations for multi-source transfer. For single-source transfer, the remaining papers often fix one dataset as a source and examine transfer to other datasets, which introduces the possibility that their results may be due to the idiosyncracies with that chosen source dataset. For multi-source transfer, it is even harder to describe the trend: each paper seems to have their own considerations for making their (non-exhaustive) choices.

These observations raise two important and related points. One: We need systematic investigations using a larger number of datasets, and two: given a large set of datasets, we need to study much more source-target combinations---ideally, we would do so exhaustively. Of course, there are practical reasons why this is difficult: even considering single-source transfer, the number of source-target combinations scales with the square of the number of datasets (for $n$ datasets, this is $(n)(n-1)$): the scaling for multi-source transfer is even worse. But think about why we need to do so: we know that datasets vary widely with respect to how the data was collected, the contexts in which the emotions are expressed, and the demographics represented in the sample. Understanding such heterogeniety is key to understanding generalization across different domains. And so we need systematic investigations into this issue if we are ever to prove that facial expression recognition is generalizable (across contexts), which is a cornerstone of scientific validity \cite{barrett2019emotional, ong2021ethical}.

This paper addresses both gaps, by using 12 datasets (a 50\% increase over the previous maximum of 8), and by systematically considering all source-target combinations for single-source transfer, and including a ``leave-one-out" design for multi-source transfer.

\section{Empirical Evaluation Details}



\begin{figure}[tbh]
\centering
\includegraphics[width=1\columnwidth]{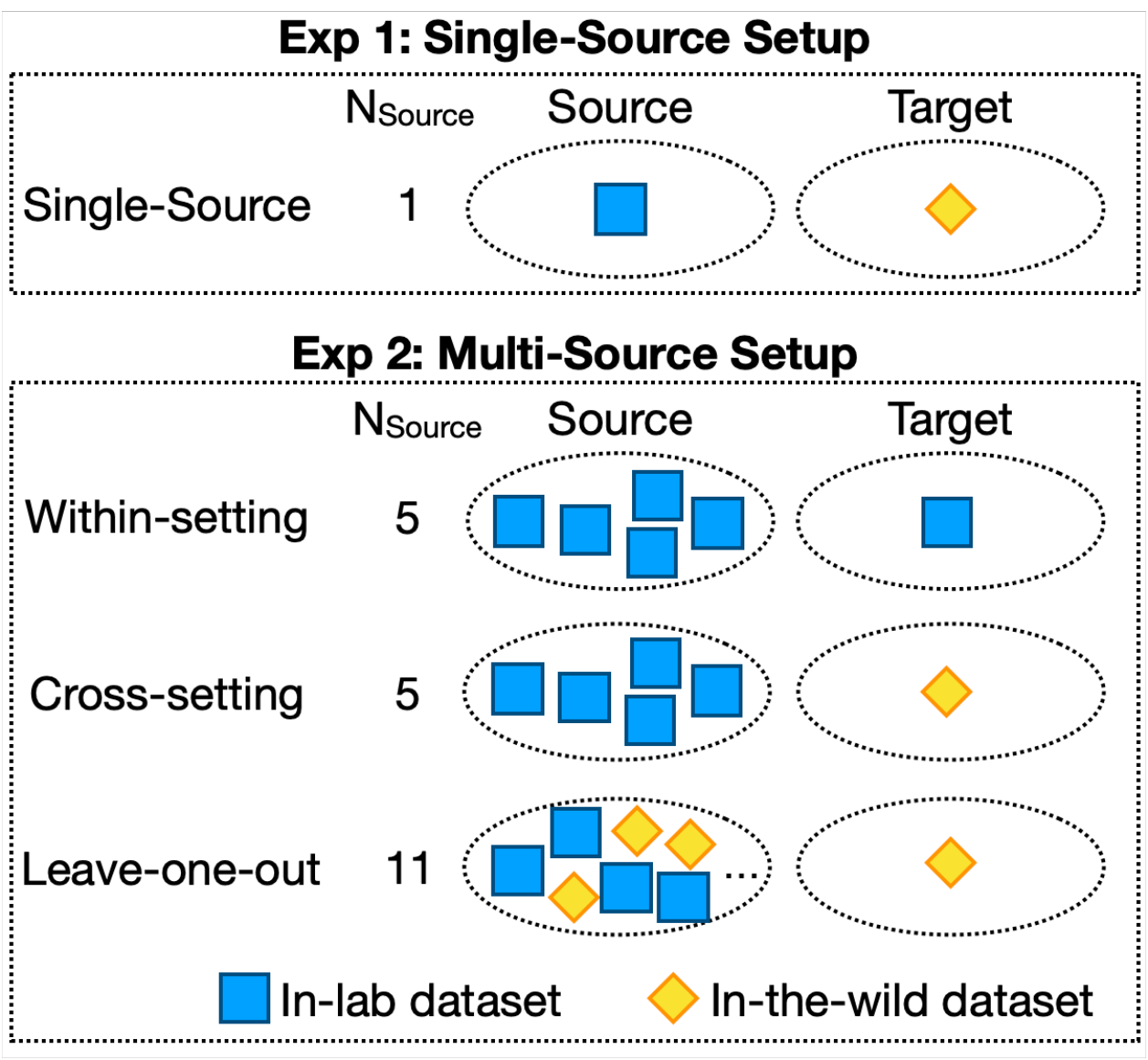}
\caption{Schematic of experiments. One example permutation is shown. Top: Experiment 1: Single-source setup, where we train the model with only one source dataset and test on other target datasets. Blue squares represent \emph{in-lab} and yellow triangles represent \emph{in-the-wild} datasets. Bottom: Experiment 2: We conducted three conditions for multi-source training. For the within-setting and cross-setting conditions, we used 5 source datasets with the same setting (i.e., 5 \emph{in-lab} or 5 \emph{in-the-wild} datasets), where the difference is only whether the target dataset comes from the same or different setting. For the Leave-one-out condition, we used the maximum number (11) of source datasets, and tested on the last target dataset.}
\label{fig:experiments}
\end{figure}

We designed three experiments to investigate the generalization of models trained on one subset of datasets, to a target dataset (Fig. \ref{fig:experiments}). In these experiments, we primarily varied the choice and combination of source datasets. We evaluated models on the test split of the target datasets. In the simplest, single-source case, \textbf{Experiment 1}, we trained a model on a single dataset. In \textbf{Experiment 2}, we trained a model using multiple source datasets.  
We performed three types of source-dataset combinations. We combined multiple datasets from same setting (i.e., either in-the-wild or in-lab) to test the generalizability within and across settings. And lastly, we combined multiple datasets from multiple settings (both in-the-wild and in-lab), in a leave-one-out manner where we use eleven source datasets. 
In \textbf{Experiment 3}, we assessed three commercial APIs, 
Amazon Rekognition, Megvii Face++, and Microsoft Azure. 

\subsection{Datasets}
\label{sec:eval_datasets}
We used six in-lab datasets: JAFFE, GEMEP, CK+, Oulu-CASIA, KDEF and, IASLab and six in-the-wild datasets: EmotioNet, SFEW, RAF-DB, Aff-Wild2, FER-2013 and AffectNet (see Fig. \ref{fig:exampleImages}). For all datasets, we use the same subset of seven classes, six emotions and neutral: \{\emph{anger}, \emph{disgust}, \emph{fear}, \emph{happiness}, \emph{sadness}, \emph{surprise}\} and \{\emph{neutral}\}. The exceptions are EmotioNet and GEMEP which do not have a \emph{neutral} class.

\begin{table*}[h!]
\centering
\resizebox{\textwidth}{!}{
\begin{tabular}{ll|ccccccc} 
\hline
\textbf{Dataset} & \textbf{Setting~} & \textbf{Neutral} & \textbf{Anger} & \textbf{Disgust} & \textbf{Fear} & \textbf{Happy}  & \textbf{Sad}  & \textbf{Surprise}  \\ 
\hline
JAFFE            & in-lab            & 24 (14.1\%)      & 24 (14.1\%)    & 23 (13.5\%)      & 25 (14.7\%)   & 25 (14.7\%)     & 25 (14.7\%)   & 24 (14.1\%)        \\
GEMEP            & in-lab            & 0                & 46 (21.7\%)    & 23 (10.8\%)      & 34 (16.0\%)   & 45 (21.2\%)     & 43 (20.3\%)   & 21 (9.9\%)         \\
CK+              & in-lab            & 98 (28.4\%)      & 36 (10.4\%)    & 47 (13.6\%)      & 20 (5.8\%)    & 55 (15.9\%)     & 23 (6.7\%)    & 66 (19.1\%)        \\
Oulu-CASIA       & in-lab            & 64 (14.3\%)      & 64 (14.3\%)    & 64 (14.3\%)      & 64 (14.3\%)   & 64 (14.3\%)     & 64 (14.3\%)   & 64 (14.3\%)        \\
KDEF             & in-lab            & 112 (14.3\%)     & 112 (14.3\%)   & 112 (14.3\%)     & 112 (14.3\%)  & 112 (14.3\%)    & 112 (14.3\%)  & 112 (14.3\%)       \\
IASLab           & in-lab            & 158 (13.7\%)     & 156 (13.5\%)   & 162 (14.1\%)     & 170 (14.8\%)  & 170 (14.8\%)    & 170 (14.8\%)  & 166 (14.4\%)       \\
EmotioNet        & in-the-wild       & 0                & 54 (5.9\%)     & 49 (5.4\%)       & 39 (4.3\%)    & 609 (66.8\%)    & 107 (11.7\%)  & 54 (5.9\%)         \\
SFEW             & in-the-wild       & 120 (16.2\%)     & 142 (19.1\%)   & 43 (5.8\%)       & 65 (8.7\%)    & 158 (21.3\%)    & 138 (18.6\%)  & 77 (10.4\%)        \\

RAF-DB            & in-the-wild       & 2019 (20.6\%)    & 564 (5.7\%)    & 574 (5.8\%)      & 225 (2.3\%)   & 3817 (38.9\%)   & 1585 (16.1\%) & 1032 (10.5\%)      \\
Aff-Wild2         & in-the-wild       & 5874 (39.4\%)    & 606 (4.1\%)    & 331 (2.2\%)      & 288 (1.9\%)   & 4040 (27.1\%)   & 2672 (17.9\%) & 1080 (7.3\%)       \\
FER-2013          & in-the-wild       & 4965 (17.3\%)    & 3995 (13.9\%)  & 436 (1.5\%)      & 4097 (14.3\%) & 7215 (25.1\%)   & 4830 (16.8\%) & 3171 (11.0\%)      \\
AffectNet        & in-the-wild       & 59899 (26.4\%)   & 19906 (8.8\%)  & 3042 (1.3\%)     & 5102 (2.2\%)  & 107532 (47.3\%) & 20367 (9.0\%) & 11272 (5.0\%)      \\
\hline
\end{tabular}
}
\vspace{0.01cm}
\caption{Distribution of emotion labels across the 12 datasets we used. We only considered these seven classes across our experiments (some datasets like AffectNet have more classes that we did not use). Numbers indicate number of samples (and as a percentage of training dataset). GEMEP and EmotioNet do not have `Neutral' images.}
\label{tab:dataset_breakdown}
\end{table*}

\begin{enumerate}

\item The JApanese Female Facial Expression (JAFFE) \cite{lyons1998coding} dataset contains of 213 images from 10 different Japanese female participants, who were asked to show facial expressions corresponding to six emotions, as well as neutral expressions. The original paper does not provide a train/validation/test split, so we created a 80/10/10 split for our experiments. 
      
\item The GEneva Multimodal Emotion Portrayals (GEMEP) \cite{banziger2012introducing} dataset is a video dataset that has 10 actors portraying 18 affective states in the lab. As the GEMEP dataset contains videos with a frame rate of 25 frames per second, we sampled a still frame for every 12 frames from all the videos. Each video is labelled with one of six emotions, and so we label frames with the label of the video. In total, we extracted 265 frames from the GEMEP dataset. We note that GEMEP does not have \emph{neutral} images. The original paper does not provide a train/validation/test split, so we created a 80/10/10 split for our experiments. 
    
\item The extended Cohn-Kanade (CK+) \cite{lucey2010extended} dataset comprises posed expression sequences (from neutral to peak expression). We select a subset of 309 sequences that are labelled with the six emotions listed above. In each sequence, we extract the final (peak) image, which is labelled with one of six emotions. In addition, one neutral image is taken from each participant. The original paper does not provide a train/validation/test split, so we created a 80/10/10 split for our experiments. 

\item The Oulu-CASIA \cite{zhao2011facial} dataset consists of image sequence (neutral to peak expression) taken from 80 subjects. Each subject has an image sequence labelled with one of six emotion categories (80*6 = 480 images). For our experiment we take the last frame of each sequence, along with one neutral image per participant. This dataset also consists of images taken in both visible and near infrared with three types of illumination, but for the purposes of our experiment (and following previous papers \cite{dias2022cross,li2020deeper}), we only consider images taken in visible light with normal illumination. The original paper does not provide a train/validation/test split, so we created a 80/10/10 split for our experiments. 
    
\item  The Karolinska Directed Emotional Faces (KDEF) \cite{lundqvist1998karolinska} dataset consists of still images from 70 amateur actors. Each image is annotated with the six emotions we are interested in along with neutral, and also taken from 5 different angles. For our experiments (and following previous papers \cite{zavarez2017cross,dias2022cross}, we use the 980 images that consists of only the frontal face images. The original paper does not provide a train/validation/test split, so we created a 80/10/10 split for our experiments. 
 
\item The Interdisciplinary Affective Science Laboratory (IASLab) dataset contains 50 unique participants who were tasked to act out a total of 9 affective states. We use a subset of 1,449 images, which correspond to the six emotions in our study, and \emph{neutral}. The original paper does not provide a train/validation/test split, so we created a 80/10/10 split for our experiments. 

\item The EmotioNet \cite{fabian2016emotionet} dataset consists of face images collected using web search engines by using keywords from WordNet starting with the root word “feeling.” A subset of this dataset is labelled with 23 emotion categories: We used the 1,140 images that belonged to one of six emotion classes in our study (EmotioNet does not contain the Neutral class). The original paper does not provide a train/validation/test split, so we created a 80/10/10 split for our experiments.

\item The Static Facial Expressions in the Wild (SFEW) \cite{dhall2012collecting} dataset consists of images extracted from the video dataset Acted Facial Expressions in the Wild (AFEW). This dataset consists of 1,365 images extracted from movies, where each image is labelled with one of the six emotions or neutral. We used the validation set provided by the authors as our held-out test set, and we divided the original training set into a 80:20 training/validation split. 

\item The Real-world Affective Faces Database (RAF-DB, \cite{li2019reliable}) dataset contains 30K images collected from flickr and filtered using keywords such as ``smile", ``shocked", and ``disgust". We used the version of the dataset annotated with the six emotions, and neutral. 
We used the test set provided by the authors, and we divided the original training set into a 80:20 training/validation split.

\item The Aff-Wild2 \cite{kollias2018aff} dataset consists of YouTube reaction videos and was released as a part of the ABAW challenge \footnote{\url{https://ibug.doc.ic.ac.uk/resources/iccv-2021-2nd-abaw/}}. In our work we use the frame images and frame-level annotations provided by the authors of the dataset \cite{kollias2018aff}. Frames are annotated with the six emotions and \emph{neutral}. These frame images were extracted from the videos that have an average frame rate of 30 frames per second and we sampled after every 30 frames. In total, we extracted 26,553 frames. We used the validation set provided by the authors as our held-out test set, and we divided the original training set into a 80:20 training/validation split. 

\item The FER-2013 \cite{goodfellow2013challenges} dataset contains 35,887 images in seven categories, obtained using Google Search from a set of 184 emotion-related keywords like ``blissful" and ``enraged". We use the train/validation/test split provided by the authors for our evaluation. 

\item The AffectNet \cite{mollahosseini2017affectnet} dataset contains facial images collected by querying 3 major search engines using 1,250 emotion related keywords in six different languages. We use a subset of 460,039 images that are labelled with the six emotions and neutral. We used the validation set provided by the author as our held-out test set, and we divided the original training set into a 80:20 training/validation split.

\end{enumerate}

The distribution of labels are given in Table \ref{tab:dataset_breakdown}, and we note that some datasets are heavily imbalanced (i.e., they do not contain an equal distribution among the emotion classes). We summarize the dataset sizes in Table \ref{tab:dataset_train_val_test}.

\begin{table}[]
\resizebox{\columnwidth}{!}{%
\begin{tabular}{l|cccc} 
\hline
\textbf{Dataset} & \textbf{Train}  & \textbf{Validation} & \textbf{Test} & \textbf{Total }  \\ 
\hline
JAFFE            & 170 (79.8\%)    & 21 (9.9\%)          & 22 (10.3\%)   & 213              \\
GEMEP            & 212 (80.0\%)    & 26 (9.8\%)          & 27 (10.2\%)   & 265              \\
CK+              & 345 (79.9\%)    & 43 (10.0\%)         & 44 (10.2\%)   & 432              \\
Oulu-CASIA       & 448 (80.0\%)    & 56 (10.0\%)         & 56 (10.0\%)   & 560              \\

KDEF             & 784 (80.0\%)    & 98 (10.0\%)         & 98 (10.0\%)   & 980              \\
IASLab           & 1152 (80.0\%)   & 144 (10.0\%)        & 144 (10.0\%)  & 1,440             \\
EmotioNet        & 912 (80.0\%)    & 114 (10.0\%)        & 114 (10.0\%)  & 1,140             \\
SFEW             & 743 (54.4\%)    & 186 (13.6\%)        & 436 (31.9\%)  & 1,365             \\

RAF-DB            & 9,816 (64.0\%)   & 2,455 (16.0\%)       & 3,068 (20.0\%) & 15,339            \\
Aff-Wild2         & 14,891 (56.1\%)  & 3,723 (14.0\%)       & 7,939 (29.9\%) & 26,553            \\
FER-2013          & 28,709 (80.0\%)  & 3,589 (10.0\%)       & 3,589 (10.0\%) & 35,887            \\
AffectNet        & 227,120 (79.0\%) & 56,781 (19.8\%)      & 3,500 (1.2\%)  & 287,401           \\
\hline
\end{tabular}
}
\vspace{0.01cm}
\caption{Number of samples in the Training / Validation / Test partitions we used. Note that these numbers are for the seven classes that we consider, so they may differ from the dataset sizes reported in the original papers.}
\label{tab:dataset_train_val_test}
\end{table}

\begin{table*}[]
\centering
\resizebox{\textwidth}{!}{
\begin{tabular}{l|cccccccccccc}
\multicolumn{1}{l|}{}    & \multicolumn{12}{c}{\textbf{Target Dataset}} \\
\textbf{Source Dataset} & JAFFE               & GEMEP               & CK+                 & Oulu-CASIA          & KDEF                & IASLab              & EmotioNet           & SFEW                & RAF-DB              & Aff-Wild2           & FER-2013            & AffectNet            \\ 
\hline
JAFFE                   & \uline{90.9 (0.0)}  & 23.8 (5.7)          & 35.0 (4.1)          & 26.1 (6.0)          & 33.7 (1.6)          & 31.9 (4.2)          & 42.5 (12.6)         & 19.7 (4.9)          & 28.9 (5.7)          & 18.7 (3.6)          & 26.5 (3.7)          & 23.4 (0.9)           \\
GEMEP                   & 17.3 (5.0)          & \uline{83.1 (3.4)}  & 17.3 (5.0)          & 23.6 (5.7)          & 25.1 (4.1)          & 22.5 (4.1)          & 31.1 (5.7)          & 24.1 (2.9)          & 25.0 (2.7)          & 12.4 (1.0)          & 24.8 (3.1)          & 20.5 (1.1)           \\
CK+                     & 24.5 (4.1)          & 15.4 (5.4)          & \uline{90.9 (3.2)}  & 49.3 (2.4)          & 54.3 (4.9)          & 51.2 (1.9)          & 53.9 (4.4)          & 26.4 (1.9)          & 49.2 (1.4)          & 38.2 (1.4)          & 37.6 (3.6)          & 33.8 (1.1)           \\
Oulu-CASIA              & 38.2 (7.6)          & 14.6 (4.2)          & \textbf{85.5 (3.4)} & \uline{65.4 (3.7)}  & 66.5 (1.8)          & 51.7 (2.0)          & 60.2 (5.6)          & 19.9 (2.5)          & 40.7 (1.6)          & 16.7 (2.5)          & 28.9 (1.8)          & 31.6 (1.1)           \\
KDEF                    & 35.5 (3.8)          & 23.1 (0.0)          & 82.3 (3.4)          & 41.8 (5.4)          & \uline{89.8 (1.9)}  & 59.0 (1.8)          & 67.2 (2.5)          & 24.6 (2.0)          & 40.0 (1.8)          & 26.5 (3.8)          & 31.6 (1.6)          & 36.2 (0.7)           \\
IASLab                  & 42.7 (7.6)          & 19.2 (5.4)          & 68.2 (4.5)          & 45.7 (4.5)          & 67.3 (2.5)          & \uline{91.2 (1.1)}  & 68.2 (1.7)          & 21.6 (2.3)          & 38.0 (2.1)          & 28.9 (4.2)          & 30.1 (2.8)          & 37.7 (1.7)           \\
EmotioNet               & 23.6 (7.5)          & 26.2 (5.0)          & 52.7 (1.9)          & 42.9 (4.2)          & 52.9 (4.0)          & 42.4 (3.0)          & \uline{89.1 (1.2)}  & 26.7 (0.4)          & 43.0 (2.1)          & 25.0 (2.6)          & 35.5 (4.3)          & 34.8 (0.9)           \\
SFEW                    & 15.5 (2.5)          & 30.8 (2.7)          & 48.2 (6.9)          & 27.1 (4.4)          & 44.1 (2.2)          & 42.4 (1.4)          & 70.5 (5.4)          & \uline{45.7 (2.0)}  & 49.5 (0.7)          & 34.4 (3.0)          & 42.0 (0.5)          & 33.3 (1.0)           \\
RAF-DB                  & \textbf{51.8 (4.1)} & 23.1 (3.8)          & 80.0 (3.7)          & 55.0 (2.6)          & 60.8 (2.1)          & 64.0 (0.9)          & 76.5 (1.9)          & 45.6 (0.7)          & \uline{85.3 (0.2)}  & 49.9 (1.6)          & 54.6 (0.7)          & \textbf{45.3 (0.4)}  \\
Aff-Wild2               & 23.6 (3.8)          & 16.2 (1.7)          & 60.0 (5.2)          & 34.3 (5.3)          & 39.8 (4.0)          & 35.8 (3.6)          & 59.5 (4.3)          & 35.2 (1.3)          & 57.6 (0.8)          & \uline{58.5 (0.8)}  & 42.9 (0.5)          & 29.5 (0.3)           \\
FER-2013                & 36.4 (3.2)          & \textbf{36.2 (2.1)} & 75.0 (3.6)          & 50.0 (3.3)          & 55.3 (1.7)          & 46.4 (4.5)          & 69.5 (3.0)          & 44.4 (1.1)          & 65.0 (0.9)          & 50.9 (1.2)          & \uline{70.5 (0.6)}  & 43.2 (0.6)           \\
AffectNet               & 45.5 (0.0)          & 24.6 (6.4)          & 83.6 (3.7)          & \textbf{60.0 (3.0)} & \textbf{77.8 (3.4)} & \textbf{80.4 (4.8)} & \textbf{82.1 (2.4)} & \textbf{53.0 (0.8)} & \textbf{79.8 (1.6)} & \textbf{58.8 (1.3)} & \textbf{57.7 (0.5)} & \uline{55.8 (1.0)}   \\ 
\hline\hline
Min. Difference      & -39.1               & -46.9               & -5.4                & -5.4                & -12                 & -10.8               & -7                  & 7.3                 & -5.5                & 0.3                 & -12.8               & -10.5                \\
Avg. Difference      & -58.7               & -60.1               & -28.4               & -24                 & -37.3               & -43.2               & -27.2               & -14.7               & -38.3               & -25.7               & -33                 & -22.2                \\
\hline
\end{tabular}
}
\vspace{0.01cm}
\caption{Experiment 1: Results of single-source, single-target experiment using ResNet50. Values indicate the average of seven-class classification accuracy over 5 runs, with standard deviation given in parentheses. Rows correspond to source datasets, and columns to target datasets. Diagonal entries (underlined) indicate the reference performance when model is trained and tested on the same dataset (i.e., within-corpus performance) while the off-diagonal entries are for cross-corpus to the target dataset. Best-performing values for each column, excluding the diagonals, are bolded. Models trained on GEMEP and EmotioNet as source did not see any \emph{neutral} class during training. The last two rows give the minimum difference in cross-corpus performance compared to within-corpus (i.e., bolded values $-$ underlined) and the averaged difference in cross-corpus performance compared to within-corpus (i.e., average of non-underlined values $-$ underlined).}
\label{tab:single_resnet}
\end{table*}

\subsection{  Model Architectures  }
\label{sec:eval_model}

We choose two model architectures: (i) ResNet-50 \cite{he2016deep} and (ii) Inception-ResNet \cite{szegedy2017inception}. These models are commonly used as base architectures and baselines for Facial Expression Recognition \cite{kollias2021analysing,li2020deeper,hasani2017spatio,chen2020stcam}. Previous research has found better facial expression recognition performance when these models are pre-trained on facial recognition datasets, as opposed to generic object recognition datasets like ImageNet. Hence, we used a ResNet-50 model pre-trained on VGGFace2 dataset \cite{cao2018vggface2} and an Inception-ResNet model pre-trained on CASIA-WebFace dataset \cite{yi2014learning}. 
And for our multi-source experiments, we also implement a recent approach proposed for domain generalization using Entropy Regularisation \cite{zhao2020domain}. We integrate entropy regularisation into a ResNet-50 model pre-trained on VGGFace2. 

Because our results were qualitatively similar across all three models, we report only the first ResNet-50 model (without domain generalization) in the main text. We report the results of the Inception-ResNet model and the ResNet-50 model with domain generalization in the Appendix.

\subsection{Implementation Details}
\label{sec:eval_implementation}

All images were first pre-processed with MTCNN \cite{zhang2016joint} to detect faces and facial landmarks. 
All images were aligned using an affine transformation with the landmark location of the eyes, and were all subsequently converted to grayscale. The images were resized to a 224 x 224 pixel square for the ResNet50 model, and 160 x 160 pixel square for the Inception-ResNet model. 
We performed horizontal flipping for a randomly selected 50\% of the images to augment training and prevent overfitting.

To train our models, we utilised a NVIDIA RTX6000 GPU. We set the learning rate to 0.001 and used a step learning rate scheduler with a step size of 10 and a decay factor $\gamma$ of 0.5. We trained our models using Stochastic Gradient Descent, with a momentum value of 0.9 for 30 epochs. We used early stopping to choose the best-performing model, based on validation accuracy. For single-source experiments, we used a batch size of 64, while for all multi-source experiments, we used a batch size of 128. For training using multiple datasets (Experiment 2), we combined and shuffled the training sets of the respective source datasets.

To evaluate our models, we report the top-1 accuracy, which is simply the proportion of correctly predicted samples. As most of the datasets are imbalanced, we also report the weighted F1-score in the Appendix.

\begin{table*}[]
\centering
\resizebox{\textwidth}{!}{
\begin{tabular}{l|cccccc||cccccc}
\multicolumn{1}{c|}{}                             & \multicolumn{12}{c}{\textbf{Target Dataset}}                         \\
\multicolumn{1}{c|}{\textbf{Source Datasets}} & JAFFE & GEMEP & CK+ & Oulu-CASIA & KDEF & IASLab & EmotioNet & SFEW & RAF-DB & Aff-Wild2 & FER-2013 & AffectNet \\ 
\hline
\textit{In-Lab Datasets ...} & \multicolumn{6}{c||}{Within-setting} & \multicolumn{6}{c}{Cross-setting} \\
... excluding JAFFE
& 48.2 (5.2)          & -                   & -                   & -                   & -                   & -                   & 78.4 (2.0)          & 29.9 (2.4)          & 49.3 (1.8)          & 31.3 (2.9)          & 39.6 (1.6)          & 43.3 (0.8)           \\
... excluding GEMEP
& -                   & 23.1 (3.8)          & -                   & -                   & -                   & -                   & 76.7 (2.7)          & 26.0 (3.3)          & 48.0 (1.2)          & 32.5 (1.9)          & 34.5 (0.4)          & 42.8 (0.9)           \\
... excluding CK+ 
& -                   & -                   & 87.7 (3.4)          & -                   & -                   & -                   & 75.8 (3.3)          & 29.0 (2.4)          & 47.2 (2.3)          & 30.1 (2.9)          & 37.1 (1.5)          & 42.1 (0.8)           \\
... excluding Oulu-CASIA
& -                   & -                   & -                   & 55.0 (1.5)          & -                   & -                   & 80.0 (0.7)          & 30.2 (2.7)          & 50.6 (1.2)          & 35.7 (3.7)          & 40.7 (1.8)          & 43.2 (0.8)           \\
... excluding KDEF
& -                   & -                   & -                   & -                   & 76.9 (2.1)          & -                   & 74.6 (2.0)          & 27.9 (2.5)          & 49.1 (1.7)          & 30.1 (2.9)          & 39.6 (1.3)          & 41.7 (0.7)           \\
... excluding IASLab 
& -                   & -                   & -                   & -                   & -                   & 66.5 (2.9)          & 72.1 (1.9)          & 31.1 (2.4)          & 49.9 (1.1)          & 26.9 (2.0)          & 37.7 (1.5)          & 40.4 (1.0)           \\
& \multicolumn{6}{c||}{} & \multicolumn{6}{c}{}
\\
\textit{In-the-wild Datasets ... }                             & \multicolumn{6}{c||}{Cross-setting}                                                                                                 & \multicolumn{6}{c}{Within-setting}                                                                                                 \\
... excluding EmotioNet
& 50.9 (3.8)          & 30.8 (7.2)          & 87.3 (3.4)          & 67.1 (2.0)          & 77.3 (2.8)          & 77.9 (2.9)          & \textbf{83.2 (1.9)} & -                   & -                   & -                   & -                   & -                    \\
... excluding SFEW
& 50.0 (3.2) & 30.0 (3.2)          & 88.2 (3.0)          & 66.4 (3.9)          & 78.0 (1.7)          & 79.0 (2.0)          & -                   & \textbf{53.5 (1.3)} & -                   & -                   & -                   & -                    \\
... excluding RAF-DB
& 47.3 (2.5)          & 34.6 (6.1)          & 86.4 (3.6)          & 58.6 (4.3)          & \textbf{78.2 (4.1)} & \textbf{80.4 (0.8)} & -                   & -                   & \textbf{78.9 (1.2)} & -                   & -                   & -                    \\
... excluding Aff-Wild2
& \textbf{50.9 (2.0)} & 33.8 (5.0)          & \textbf{90.5 (3.0)} & \textbf{67.5 (2.6)} & 77.1 (2.7)          & 74.6 (1.4)          & -                   & -                   & -                   & \textbf{60.6 (2.3)} & -                   & -                    \\
... excluding FER-2013
& 50.0 (4.5)          & \textbf{36.2 (4.4)} & 86.4 (4.3)          & 63.6 (4.5)          & 77.1 (0.6)          & 76.9 (1.5)          & -                   & -                   & -                   & -                   & \textbf{58.7 (0.3)} & -                    \\
... excluding AffectNet
& 40.0 (8.7)          & 35.4 (3.2)          & 78.2 (6.7)          & 58.2 (1.6)          & 64.7 (4.6)          & 65.8 (3.1)          & -                   & -                   & -                   & -                   & -                   & \textbf{48.0 (0.6)}  \\ 
\hline
Within-corpus performance
& \multirow{2}{*}{90.9 (0.0)}          & \multirow{2}{*}{83.1 (3.1)}          & \multirow{2}{*}{90.9 (2.9)}          & \multirow{2}{*}{65.4 (3.3)}          & \multirow{2}{*}{89.8 (1.7)}          & \multirow{2}{*}{91.2 (0.9)}          & \multirow{2}{*}{89.1 (1.1)}          & \multirow{2}{*}{45.7 (1.7)}          & \multirow{2}{*}{85.3 (0.2)}          & \multirow{2}{*}{58.5 (0.7)}          & \multirow{2}{*}{70.5 (0.5)}          & \multirow{2}{*}{55.8 (0.9)}           \\
\;\; from Exp. 1 & & & & & & & & & & & & \\
Minimum difference                                & -40                 & -46.9               & -0.4                & 2.1                 & -11.6               & -10.8               & -5.9                  & 7.8                 & -6.4                & 2.1                 & -11.8               & -7.8                 \\
\hline
\end{tabular}
}
\vspace{0.01cm}
\caption{Experiment 2: Results of the Within-Setting and Cross-Setting conditions using ResNet50 pre-trained on VGGFace2, where each row indicates a model that is trained on five datasets from the same setting (i.e., five in-lab or five in-the-wild datasets). For example, the first row gives the model trained on five in-lab datasets (excluding JAFFE), and we have one performance value for ``within-setting" (i.e., tested on JAFFE) and six values for ``cross-setting" (i.e., tested on the six in-the-wild datasets).
Values indicate the average of seven-class classification accuracy over 5 runs, with standard deviation given in parentheses. 
Best-performing values for each column, are bolded. The last two rows give the non-transfer reference performance (from Experiment 1, i.e., the diagonal values from Table \ref{tab:single_resnet}) and the difference compared with the best-performing (bold) values.
}
\label{tab:resnet_within_cross_setting}
\end{table*}

\section{Results}

\subsection{Experiment 1: Single-Source Generalization}

In Experiment 1, we test whether models trained on one dataset generalize well to other datasets. The top-1 accuracies of the ResNet-50 model on every source-target combination are given in Table \ref{tab:single_resnet}, with the \emph{source} datasets on the rows and the \emph{target} datasets on the columns. The diagonal entries give the \textbf{within-corpus} performance when the model is trained on the training partition and evaluated on the test partition of the same dataset---these values serve as the reference. The off-diagonal entries give the cross-corpus performance where the model is trained and validated on a given source dataset's train/validation partition (rows) and evaluated on a target dataset's test partition (columns). We reiterate that the off-diagonal values show the performance of the model when predicting labels on the target dataset without seeing any target samples during training. 


First, as a sanity check, we observed that the within-corpus performance for all twelve datasets, with two exceptions (discussed below), is the highest in each column---that is, the maximum performance in each column is the diagonal entry---which is not surprising as the evaluation data comes from the same distribution as the training data. We also note that the within-corpus performances of the in-lab datasets are relatively high (from 83.1 to 91.2\%, with Oulu-CASIA being an outlier at 65.4\%), and are higher than the corresponding performances of the in-the-wild datasets (55.8\% to 89.1\%, with SFEW being an outlier at 45.7\%). We note that the aim of this paper is to assess the domain generalizability of a commonly-used model, instead of innovating on model architecture to surpass state-of-the-art performance: that said, our model performs comparably to 
other papers that use similar architectures (e.g., if we focus on the more challenging naturalistic datasets: 
AffectNet, \textit{ours:} ~58.6, \textit{reported:} 52\%-58.4\%\cite{ngo2019facial,li2020deeper}; Aff-Wild2: \textbf{ours: 57.7\%} \textit{reported: 50\%-56\%} \cite{do2020affective}). (Later, we also compare all our results against the literature, in Table \ref{tab:lit_review_results}).

Our first result is that cross-corpus performance is generally poor compared to the within-corpus model performance. The last two rows in Table \ref{tab:single_resnet} give the minimum and average drop in cross-corpus performance compared to the within-corpus model. Even if we consider only the \emph{minimum} drop in performance---i.e., the difference between the \emph{best-performing} cross-corpus model for that dataset (column) and the within-corpus model--- we see differences in performance of approximately -12.3\% on average, with the highest being a 46.9\% drop in classification accuracy (where GEMEP is the target dataset). The only exceptions are (i) the model trained on AffectNet and tested on SFEW, which outperformed the within-corpus performance by 7.3 percentage points, and (ii) the model trained on AffectNet and tested on Aff-Wild2, which differs from the within-corpus performance by a small 0.3 percentage points. 

However, taking the best-performing single-source cross-corpus model may be too generous a comparison, as we would require oracular knowledge of which would be the best-performing model in the first place. If we instead look at the \emph{average} change in performance, we see something more sobering: the mean classification accuracy across all twelve target datasets is only 42.0\% (compared to chance at 14.3\%), which translates to an average decrease in classification accuracy, compared to the within-corpus performance, of -34.4 percentage points (with a range of decreases from -14.7 to -60.1 percentage points).
We notice that 8 of the 12 best-performing cross-corpus performance (i.e., the bolded values) come from the same source dataset: AffectNet --- we can see this in the last row of Table \ref{tab:resnet_within_cross_setting}. AffectNet also happens to be the largest dataset, so perhaps one interim observation is that we can improve cross-corpus performance by increasing the number of training samples. 

We also note that there exists substantial heterogeneity in the results, which suggests that some datasets may be more idiosyncratic. For example, the classification accuracies, including the within-corpus accuracy, on Oulu-CASIA are lower compared to the other in-lab datasets, and the performances on SFEW are lower compared to the other in-the-wild datasets. Other datasets have high within-corpus accuracy, but are significantly more difficult for other cross-corpus models to transfer \emph{to}, such as JAFFE and GEMEP. 
%
%
This analysis highlights the limitations of drawing inferences about transfer learning when using a single source dataset (or small number of source datasets), as each dataset has its own idiosyncrasies (see most studies in Table \ref{tab:lit_review}).


\begin{figure*}[tbh]
\centering
\includegraphics[width=.9\textwidth]{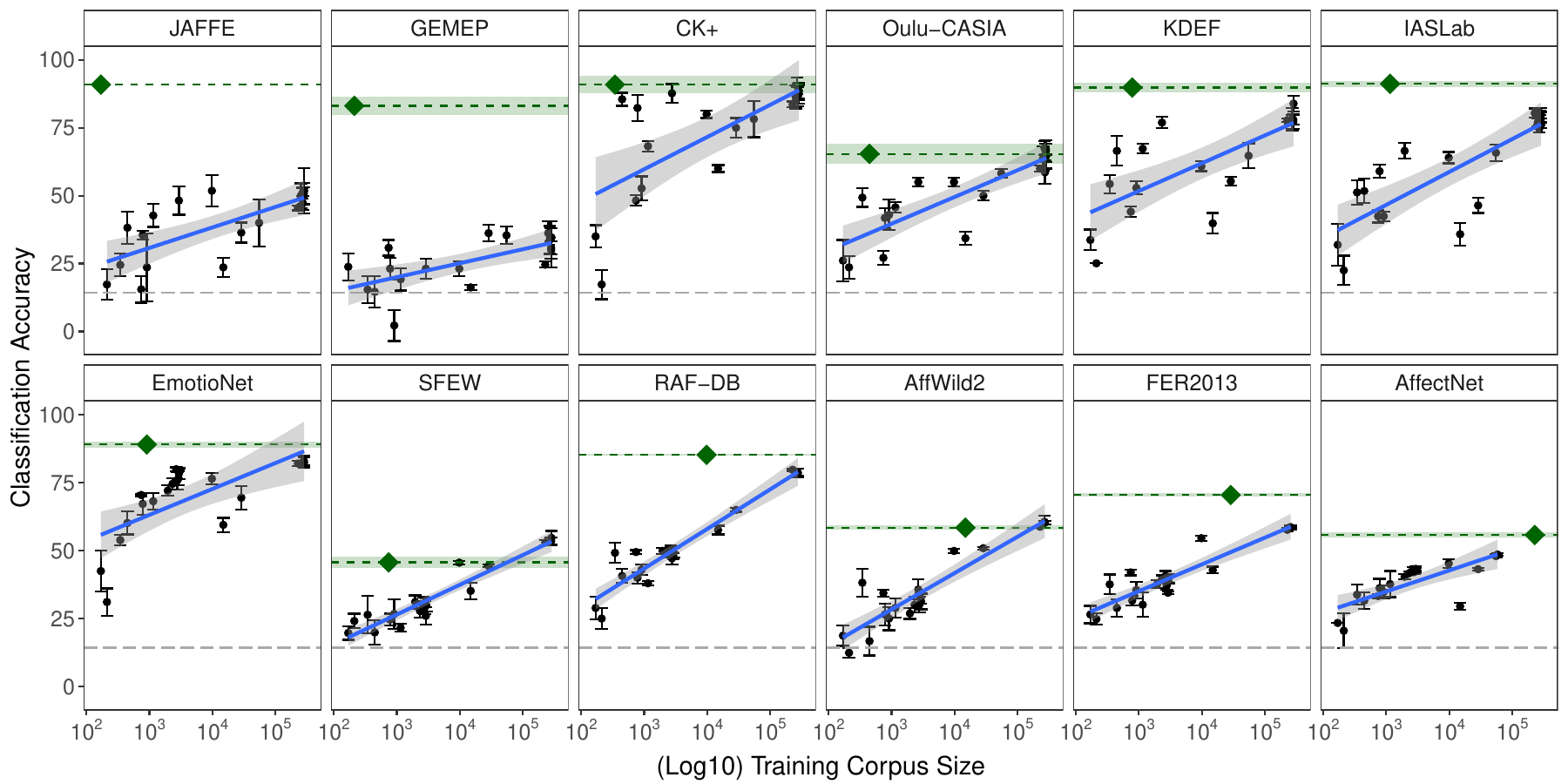}
\caption{Plots of cross-corpus classification accuracy on the vertical axis against training corpus size on the horizontal axis, by target dataset. Data-points are from Experiments 1 and 2, with standard deviation error bars. Solid green diamonds and dashed lines give the within-corpus performance as reference. Grey dashed lines indicate chance for a 7-class classification. Top row: In-lab datasets. Bottom row: In-the-wild datasets.}
\label{fig:corpus_size_plot}
\end{figure*}

\subsection{    Experiment 2: Multi-Source Generalization  }

Experiment 1 examined training on only one source dataset. Next, Experiment 2 focuses on increasing the number of source datasets. This could help the model learn domain-invariant features from different distributions. 

\subsubsection{Within- and cross-setting conditions}

In the first two conditions, we were interested in testing the intuition that the setting in which the dataset is collected matters. Researchers have traditionally distinguished posed expressions (by actors or na\"ive volunteers) that are collected \textit{in the lab} from naturalistic, spontaneous expressions collected \textit{in the wild}. Presumably, because images from in-lab datasets are more controlled and are visually similar, we might expect higher performance when a model trained on in-lab data is tested on other in-lab data (``within-setting"), than when tested on in-the-wild data (``cross-setting"). And similarly for models trained on in-the-wild data.


As we have six \emph{in-lab} and six \emph{in-the-wild} datasets, we chose to maximize the number of source datasets that would still allow us to test this hypothesis. Thus, we chose to fix the number of source datasets at five.
Specifically, we would train our model on five in-lab datasets; we can then evaluate it on the last in-lab dataset (to give us a value for within-setting cross-corpus transfer), on all six in-the-wild datasets (to give us six values for cross-setting cross-corpus transfer). We can do the same for the other in-lab datasets; as well as for all the in-the-wild datasets. 


These results are given in Table \ref{tab:resnet_within_cross_setting}.
All of the best-performing combinations (i.e., bold values in Table \ref{tab:resnet_within_cross_setting}) were obtained from the models trained using in-the-wild datasets, irrespective of the evaluation type (i.e., whether they were evaluated in within-setting cross-setting). This is somewhat surprising given the strong distinction between ``in-lab" and ``in-the-wild" datasets: one would expect that training on data collected in more similar settings would yield better performance due to the the similarity in training/test data distributions. However, there is a confound in that all the in-the-wild are larger than the in-lab datasets. The results is thus consistent with an alternate hypothesis, that increasing the number of training samples, irrespective of their setting, yields better performance. 

Compared to the best single-source transfer, we note that the best 5-source multi-source performance improves or remains the same for ten of the twelve datasets, except for a slight dip in performance for JAFFE (51.8 to 50.9) and RAF-DB (79.8 to 78.9). But when we then compare to the within-corpus performance (last row of Table \ref{tab:resnet_within_cross_setting}), we see that again for ten of the twelve datasets, the 5-source performance still underperforms the within-corpus performance (except for Oulu-CASIA and Aff-Wild2).

\subsubsection{Leave-one-out: Maximum number of source datasets}

In this condition we wanted to see what cross-corpus performance the model could achieve with the maximum amount of training data we had. For each target dataset, we train on the remaining 11 source datasets and evaluate the performance on this target dataset.
The results are given in Table \ref{tab:resnet_leave_one_out} (with a full table for the additional two models provided in the Appendix). 

%
For ten of the twelve datasets, the leave-one-out accuracy still underperforms the within-corpus classification performance. Averaging across all twelve datasets, the mean seven-class classification accuracy is 65.6\%, which corresponds to a mean performance drop of 10.8 percentage points, compared to the within-corpus classification. 
In general, we see that this experiment also corroborates the finding from the previous two conditions, that the larger the amount of data generally improves the cross-corpus performance for all datasets. In order to see this more clearly, we examine these trends statistically in the next analysis. 

\begin{table}[tbh]
\centering
\begin{tabular}{l|cccc}
%
Target Dataset & Leave-one-out & Difference with within-corpus \\
\hline
JAFFE & 51.8 (8.3) & -39.1 \\
GEMEP & 34.6 (6.1) & -48.5 \\
CK+ & 88.6 (2.8) & -2.3 \\
Oulu-CASIA & 64.6 (4.6) & -0.8 \\
KDEF & 83.9 (2.8) & -5.9 \\
IASLab & 80.4 (2.0) & -10.8 \\
EmotioNet & 82.5 (1.8) & -6.6 \\
SFEW & 54.7 (2.6) & 9.0 \\
RAF-DB & 78.6 (1.6) & -6.7 \\
Aff-Wild2 & 60.3 (0.6) & 1.8 \\
FER-2013 & 58.5 (0.7) & -12.0 \\
AffectNet & 48.6 (0.5) & -7.2 
\end{tabular}
\vspace{0.01cm}
\caption{Experiment 2: Results of the Leave-one-out condition using ResNet50 pre-trained on VGGFace2. In this condition we train on 11 source datasets and report the classification accuracies on the test partition of the last, held-out test set. We also report the differences in classification accuracies compared with the within-corpus results from Experiment 1. 
Negative values indicate that the leave-one-out condition performed worse than the within-corpus.
}
\label{tab:resnet_leave_one_out}
\end{table}

\begin{table*}
\centering
\resizebox{\textwidth}{!}{
\begin{tabular}{l|c|c|c|c|c}
\textbf{Dataset} & \textbf{Within-corpus} & \textbf{Cross-corpus} & \textbf{None}                                                                                                       & \textbf{Generalization}                                                                 & \textbf{Adaptation}                                             \\
                 & (Ours)                 & (Ours) &          &  & \\ \hline \hline
JAFFE            & 90.9                   & 51.8                  & \begin{tabular}[c]{@{}c@{}}22.02 \cite{zhu2015atransfer}, 37.36 \cite{lopes2017facial}, 40.98 \cite{li2020facial}, 
\\41.3 \cite{shan2009facial}, 42.3 \cite{da2015effects}, 44.32 \cite{zavarez2017cross}, 
\\48.67 \cite{ali2016boosted}, 50.23 \cite{yang2020effects}, 50.7 \cite{wen2017ensemble}, 
\\55.87 \cite{gu2012facial}\end{tabular}                                                                                            & 48.13 \cite{wang2019cross}    & \begin{tabular}[c]{@{}c@{}} 57.75 \cite{li2018deep}, 58.51 \cite{miao2012cross}, 61.5 \cite{chen2021cross}, 
\\61.94 \cite{li2020deeper}, 63.38 \cite{zhu2016discriminative}, 68.54 \cite{li2021jdman}, \\74.07 \cite{zhou2020uncertainty}  \end{tabular}  \\ \hline
GEMEP            & 83.1                   & 34.6                  & -                                                                 & -                                                                                       & -  \\ \hline                                                                                          
CK+              & 90.9                   & 88.6                  & \begin{tabular}[c]{@{}c@{}}38.57 
\cite{yang2020effects}, 46.84 \cite{zhu2015atransfer}, 54.05 \cite{gu2012facial}, 
\\57.6 \cite{da2015effects}, 61.2 \cite{zhang2015facial}, 62.0 \cite{li2017reliable}, 
\\64.2 \cite{mollahosseini2016going}, 64.4 \cite{li2020facial}, 66.2 \cite{mayer2014cross}, 
\\71.29 \cite{meng2017identity}, 73.9 \cite{hasani2017spatio}, 76.05 \cite{wen2017ensemble}, 
\\84.47 \cite{xie2019sparse}, 85.8 \cite{cruz2014one}, 88.58 \cite{zavarez2017cross}, 
\\91.67 \cite{zeng2018facial}, 93.46 \cite{liu2015inspired}\end{tabular} & 85.75 \cite{wang2019cross}, 88.7 \cite{ji2019cross}   & 

\begin{tabular}[c]{@{}c@{}} 78.83 \cite{li2018deep}, 84.42 \cite{zhou2020uncertainty}, 85.71 \cite{chen2021cross}, \\88.51 \cite{li2021jdman}, 89.69 \cite{li2020deeper} \end{tabular}\\ \hline
Oulu-CASIA       & 65.4  & 64.6  & 50.83 \cite{xie2019sparse}, 61.02 \cite{zeng2018facial}  & -  & 63.97 \cite{li2020deeper}, 64.38 \cite{li2021jdman}  \\ \hline 
KDEF             & 89.8                   & 83.9                  & 72.55 \cite{zavarez2017cross} & 76.45 \cite{wang2019cross}, 84.9 \cite{dias2022cross} & - \\ \hline
IASLab           & 91.2                   & 80.4                  & -                                                               & -                                                               & -      \\ \hline                                                       
EmotioNet        & 89.1                   & 82.5                  & -                                                               & 62.3 \cite{ji2019cross}                                                & -      \\ \hline
SFEW             & 45.7                   & 54.7                  & 29.43 \cite{liu2015inspired}, 39.8 \cite{mollahosseini2016going}, 58.29 \cite{zeng2018facial}        & 49.4 \cite{ji2019cross} & \begin{tabular}[c]{@{}c@{}} 47.55 \cite{li2018deep}, 51.7 \cite{yan2019cross}, 53.21 \cite{zhou2020uncertainty}, \\ 54.34 \cite{li2020deeper}, 56.43 \cite{chen2021cross}  \end{tabular} \\ \hline
RAF-DB           & 85.3                   & 78.6                  & 39.0 \cite{li2017reliable}                                                                        & 43.8 \cite{ji2019cross}    & 53.1 \cite{yan2019cross} \\ \hline
Aff-Wild2        & 58.5                   & 60.3                  & -                                                                                                                                             & -    & -   \\ \hline
FER-2013  & 70.5  & 58.5  & 34.0 \cite{mollahosseini2016going}, 60.19 \cite{xie2019sparse}  & -   & \begin{tabular}[c]{@{}c@{}} 52.37 \cite{li2018deep}, 58.21 \cite{li2020deeper}, 58.63 \cite{li2021jdman}, \\ 58.95 \cite{chen2021cross}   \end{tabular}\\ \hline
AffectNet        & 55.8                   & 48.6                  & -                                                               & & 51.84 \cite{li2020deeper}, 52.54 \cite{li2021jdman}  \\
\hline
\end{tabular}
}
\vspace{0.1cm}
\caption{Summary of cross-corpus facial expression recognition performance from the literature. Values indicate classification accuracies. For the second and third columns, we reproduce the within-corpus results from Experiment 1, and the cross-corpus leave-one-out performance from Experiment 2. 
For each of the papers (with their citations in square brackets), we report the performance corresponding to the best-performing model they report in their paper. 
Like Table \ref{tab:lit_review}, we group papers into three groups: if they did not incorporate any  domain generalization or adaptaion mechanism (``None"), if they incorporated domain generalization techniques (``Generalization") or if they incorporated domain adaptation techniques (``Adaptation"). Adaptation approaches use a small amount of target samples (as compared to seeing no target samples at all for None and Generalization, as well as our cross-corpus results), and hence we would expect the Adaptation results to be higher.
}
\label{tab:lit_review_results}
\end{table*}

\subsubsection{Statistically testing the effect of dataset size}

The results of Experiment 2 suggest that the size of the training corpus seems to be an important factor that affects cross-corpus performance (i.e. greater the size of the training corpus, the better the cross-corpus performance), and perhaps more so than the hypothesis we started with about the setting of the datasets (in-lab or in-the-wild). In order to statistically verify the dependence of the cross-corpus classification performance with the size of the training corpus, we ran statistical analyses using mixed-effects linear models, attempting to predict classification accuracy (as the dependent variable) using corpus size. As dataset sizes varied over orders of magnitude, we converted the sizes of the training corpus to its base-10 logarithm. 

We obtained all the cross-corpus performance for each target dataset from Experiments 1 and 2. For each target dataset, we have 19 data points---From Experiment 1, 11 data points for single-source cross-corpus prediction, and from Experiment 2, 1 data point for the ``within-setting" condition, 6 data points for the ``cross-setting" condition, and 1 data point for the leave-one-out condition. 

We note several characteristics of this data: One, there is idiosyncratic variation across target datasets that we would like to model (e.g., some datasets are more difficult than others). Two, individual-cross corpus performances are nested within a specific target dataset (e.g., all of the cross-corpus results on KDEF share that commonality with each other, but not with the other data points). Thus, we used a mixed-effects linear model controlling for random intercepts and random slopes nested within target dataset.

We found a statistically significant association of corpus size with performance accuracy on the target dataset ($b = 10.2$ [95\% Confidence Interval $= 8.72, 11.7$], $t=13.6$, $p<.001$). We can interpret this coefficient as suggesting that, on average across the datasets (and assuming a linear trend), that every order of magnitude increase in the training corpus size translates to a 10.2 percentage-point increase in accuracy on the target dataset.

See Figure \ref{fig:corpus_size_plot} for a visual illustration of the relationship between training corpus size and cross-corpus classification performance across all the datasets. We note two positive and two sobering observations. First, we can observe the results of the statistical model: how, for all the target datasets, increasing the training corpus size improves cross-corpus classification accuracy---even when this particular model has no specific domain-generalization mechanism. Second, for most of the target datasets, the performance of the models trained on the largest corpus sizes approaches the within-corpus performance. For some, like SFEW and AffWild2, the best cross-corpus performance can match or slightly surpass the within-corpus performance. But we also note that for others like JAFFE and GEMEP and to a smaller extent, some in-the-wild datasets like FER2013 and AffectNet, there still exists large performance gaps. The first of the two sobering conclusions is that the raw magnitudes of the classification accuracies are not very high, especially for some of the in-the-wild datasets. For SFEW, even though the cross-corpus classification accuracy for the largest corpus size matches and even slightly surpasses the within-corpus accuracy, both these accuracies are still only in the 50\% range for a 7-class classification. Finally, there also exists substantial heterogeniety across datasets. This makes it even more important for future papers to describe their training datasets in great detail, as well as to conduct systematic investigations into how their models transfer to different populations.

\begin{figure*}[tbh]
\centering
\includegraphics[width=.9\textwidth]{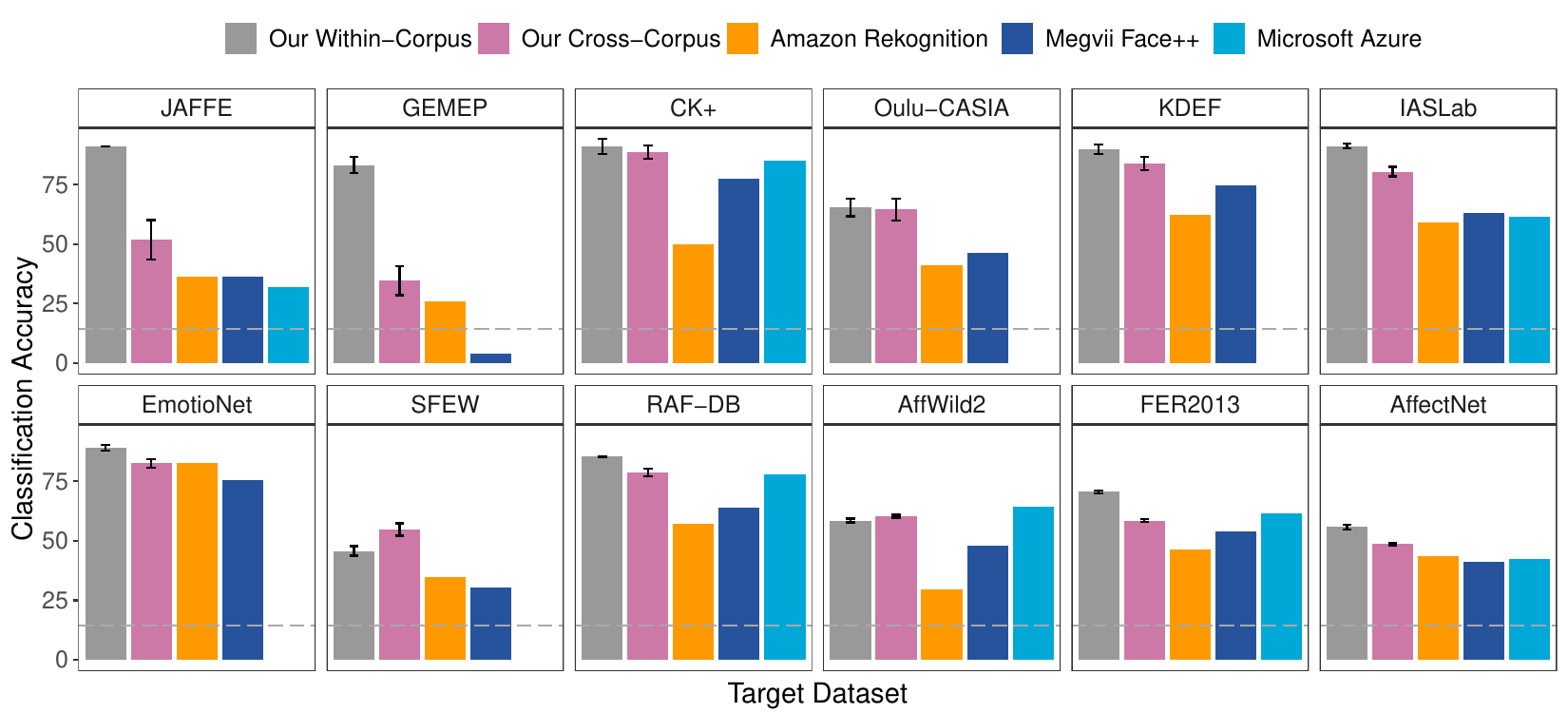}
\caption{Bar graphs showing the classification accuracies of the three commercial APIs tested, arranged by target dataset. The within-corpus and (leave-one-out) cross-corpus results from Experiments 1 and 2 are also reproduced for comparison. Grey dashed lines indicate chance for a 7-class classification. Top row: In-lab datasets. Bottom row: In-the-wild datasets.}
\label{fig:api_summary}
\end{figure*}

\subsubsection{Comparing Experiment 2 results to the literature}
We end this section with a comparison of our results with results that were previously reported in the literature and which we earlier summarized in Table \ref{tab:lit_review}. In Table \ref{tab:lit_review_results}, we document the cross-corpus results reported by each paper. In providing this table, we acknowledge that there is a lot of heterogeneity that we are aggregating out: each paper uses different models with different training protocols and with different source datasets. For example, for the most well-studied example, CK+, the 24 classification accuracies from the literature range from 38.6\% to 93.5\%. 

Table \ref{tab:lit_review_results} demonstrates that our results are not a uniquely poor characterization of corss-corpus performance in facial expression recognition. In fact, looking across each row of Table \ref{tab:lit_review_results} (corresponding to each dataset), our results are all near the higher end of the results reported in the literature.


\subsection{Experiment 3: Using Commercial APIs}

Finally, we wanted to evaluate how commerical APIs would do, on the set of datasets that we had collected. Dupr\'e and colleagues \cite{dupre2020performance} recently compared eight commercial products with human ratings collected on two databases, one posed (BU-4DFE) and one spontaneous (UT-Dallas). They considered classification of the same six emotions, but did not include neutral. They found that the performance of the automatic classifiers ranged from 48\% to 62\%, with the classifiers doing better on the posed, compared to the spontaneous expressions. (An interesting result they found is that human observers achieved a classification accuracy of 72.5\%, which is higher than the commercial products, but perhaps not as high as one might have expected.)

\begin{table}
\centering
    \resizebox{\columnwidth}{!} {
    \begin{tabular}{lc|cc|c}
    & \multicolumn{1}{c|}{Face++} & \multicolumn{2}{c}{Amazon Rekognition} & \multicolumn{1}{c}{Microsoft Azure} \\
    Dataset             & Undetected      & Undetected & Neutral  & Undetected  \\
    \hline
    \textbf{JAFFE}     & 0.0 & 0.0 & 13.6 & 0.0  \\
    \textbf{GEMEP}     & 3.7 & 3.7 & 0.0 & - \\
    \textbf{CK+}    & 0.0 & 0.0 & 29.5 & 0.0 \\
    \textbf{Oulu-CASIA}     & 0.0 & 0.0 & 14.3 & - \\
    \textbf{KDEF}     & 0.0  & 0.0 & 14.3 & - \\
    \textbf{IASLab}     & 0.0 & 0.0 & 13.9 & 0.0 \\    
    \textbf{EmotioNet}  & 0.0 & 0.0 & 0.0 & - \\
    \textbf{SFEW}  & 0.0 & 0.7 & 19.7 & - \\
    \textbf{RAF-DB} & 5.6 & 0.2 & 22.2 & 14.8 \\ 
    \textbf{Aff-Wild2} & 5.2 & 2.7 & 41.0 & 11.6 \\    
    \textbf{FER-2013} & 67.9 & 0.8 & 17.4 & 13.9 \\  
    \textbf{AffectNet} & 0.1 & 0.0 & 14.3 & 1.1 \\      
    \end{tabular}
    }
    \caption{Experiment 3: Percentage of test set that the commercial APIs could not classify. `Undetected' indicates that the API failed to detect a face.  'Neutral' indicates the percentages of the \emph{neutral} examples in the test sets, as Amazon Rekognition does not predict the \emph{neutral} class. `-' indicates that we do not have Microsoft Azure labels on those datasets.}
    \label{tab:descript_api}
\end{table}


In Experiment 3, we evaluated the generalization of three commercially-available APIs: Face++ \cite{faceplus}, Amazon Rekognition \cite{rekognition}, and Microsoft Azure \cite{azure}. The use of such APIs spans a wide range of applications in domains such as market research, healthcare, education, and finance \cite{dupre2018accuracy}. Aside from facial expressions, these APIs can also ``detect" other attributes, such as age and gender, presence of facial hair and eye glasses, and facial landmarks. We focus only on emotion attributes returned by the APIs. 

The three APIs classify facial images into a set of emotion classes along with the confidence level for each emotion. We take the emotion class with the highest confidence as that API's prediction. For Face++, the emotions classes returned map exactly to the six emotions plus \emph{neutral} that we have considered in this paper. Amazon Rekognition returns a slightly different set of emotions (nine classes), namely \{\emph{anger}, \emph{calm}, \emph{confused}, \emph{disgust}, \emph{fear}, \emph{happy}, \emph{sad}, \emph{surprised}, and \emph{unknown}\}. The six emotions we considered are a subset of the nine classes that Rekognition returns, but Rekognition does not have a \emph{neutral} class. For the case when the Rekognition API returns a result with the highest confidence level on \emph{calm}, \emph{confused}, or \emph{unknown}, we took the emotion with the next highest confidence level that is in our set as the label for that image (we note that empirically, none of the images were labelled by Rekognition as \emph{unknown}). Azure predicts the six emotions we have considered in this paper plus \emph{neutral} and additionally \emph{contempt}. If Azure returns a result with the highest confidence level on \emph{contempt}, we took the emotion with the next highest confidence level as the label for that image. On 21 June 2022, Microsoft Azure discontinued their emotion recognition API service (discussed more in the Ethical Impact Statement below), and so we only have results for seven of the twelve datasets.


We used only the test-set images of each dataset as input into the APIs. In Table \ref{tab:descript_api}, we report the percentage of images on which the APIs did not detect a face, and hence could not predict the emotion. For many of the datasets, this non-detection rate was 0\% or low, with one outlier: For an unknown reason, Face++ was unable to detect 67.9\% of the images on FER-2013 (the corresponding rates for Azure and for Rekognition were 13.9\% and 0.8\%).

Figure \ref{fig:api_summary} presents the performance of the commercial APIs. Generally, the API performances are quite similar for some datasets (e.g., JAFFE, IASLab, AffectNet), and seem to follow another pattern where performance increases from Rekognition to Face++ to Azure (e.g., CK+, RAF-DB, AffWild2, FER2013). However, we do not want to read too much into the numerical reuslts as these APIs are being updated all the time \cite{chen2021did} (or even being discontinued like Azure), and there were limitations (e.g., mismatch in the set of emotion classes, especially the lack of \emph{neutral} for Rekognition). But across all the datasets, the API performances still underperform the within-corpus results by an average of roughly 25 percentage points (For Rekognition, mean: -30\emph{pp.}, range: [-57, -7]; for Face++, mean: -25\emph{pp.}, range: [-79, -11] ; for Azure, mean: -17\emph{pp.}, range: [-59, +6]).

\section{  Discussion   }


In this study, we conducted the largest-to-date systematic investigation of domain generalization (cross-corpus prediction) using 12 facial expression recognition datasets and a comprehensive suite of experiments. The results are sobering. Generally, we conclude that single-source, single-target generalization performance for facial expression recognition is poor, achieving a mean accuracy of 42.0\% for a seven-class classification. This translates to a mean decrease in classification accuracy of 34.4 percentage points (Range: -14 to -60\emph{pp.}) compared to the within-corpus performance (mean: 76.4\%). This by itself may not be surprising given how idiosyncractic individual datasets may be. But even if we pooled data from multiple datasets together (in Experiment 2), we still find that the best-performing models still achieve an average accuracy of 65.6\%, which corresponds to a mean performance drop of 10.8 percentage points. And even commercial APIs today underperform the within-corpus results by on average 25 percentage points.

We discuss the specific takeaways from our experiments with respect to training these classifiers. Then we discuss some limitations of our experiments in particular, and limitations of the datasets and the state of the FER field more broadly. Finally, we discuss broader implications of the work, especially in light of recent psychological evidence.

First, we show that modern deep learning models perform better when trained with larger amount of data from different domains, even without specific domain generalization mechanisms. 
%
We verified across all our experiments that increasing the size of the training corpus improves cross-corpus classification accuracy, but with diminishing returns: every order of magnitude increase in corpus size translates to a 10 percentage point increase, on average.

Our results also highlight the dangers of relying on a single or small number of target-datasets for making claims about generalization. Some of the datasets in our study did not transfer well to other datasets, with very low transfer performance and/or very variable performance. We specifically chose to do a ``round-robin" format and to discuss general trends, rather than focusing on specific datasets. Moving forward, affective computing research should prioritize the need to evaluate their models on a range of different datasets, in order to validate their generalizability.

Our study, though systematic and rigorous, has several limitations. First, although we examined twelve datasets---the largest such examination in domain adaptation and generalization in facial expression recognition---there were many other datasets that we were not able to include in this study, for numerous reasons, such as they were labelled with different emotion categories than those we were interested in. Second, facial expression recognition datasets tend to have imbalanced emotion label distributions. Of the twelve datasets we examined, IASLAB was the most well balanced (with each of seven classes making up between 13.7-14.8\% of the dataset), while others like AffectNet were severely imbalanced (47.4\% Happy and 26.4\% Neutral, with the remaining 26.2\% divided among 5 classes). These dataset imbalances are known to introduce bias and impact the robustness of the models trained on these datasets, and this is a major concern especially if these or similarly-imbalanced datasets provide the training data for commercially-deployed models. Future work could examine other state-of-the-art facial expression recognition models to assess their generalizability. It could be the case that adding more datasets, having more balanced datasets, or trying different model architectures / training protocols could change several of the quantitative results we found in this study. However, we believe that addressing these limitations will lead to \emph{differences in degree}, but not \emph{differences of kind}, and hence our generic conclusions should generalize.

One huge limitation of the entire endeavour of facial expression recognition (beyond our study) is that it is confined to a small set of emotion categories, and neglects variation across cultures and demographic groups. This extensive study with so many datapoints was possible only when restricted to a small set of six emotions (plus neutral): once we consider some of the myriad other emotions that we experience in daily life, there is far fewer data out there. 

But recent psychological evidence should also challenge the field to reconsider the validity of the notion of ``facial expression recognition" \cite{barrett2019emotional}. In our opinion, the terminology suggests an equivalence to ``object recognition", but the two tasks are not close. While there is an objective truth to an object label (``cat"), and perhaps there could also be an objective label of facial movements like ``frown", it is far more complicated to attribute an emotional state to the person, or to perceive an emotion in that person. Emotions arise from subjective responses to events in the world, and are affected by a host of contextual factors, including the social and cultural settings. We should consider facial expressions as only one cue from which we have to solve this underdetermined inference problem of inferring what someone is feeling \cite{ong2015affective}. When we conceptualize the problem in this way, it naturally guides us to think about context, or multimodal inference, or considering the temporal history preceding the emotional event, or a host of other factors that suggest a more complete understanding of (perceiving) someone's emotional state. And it suggests a reason as to why ``emotion recognition" from decontextualised faces alone may even prove to be a fools' errand (and why even with today's technologies, this is not yet a ``solved problem"). 


\section{   Ethical Impact Statement   }

The proliferation of commercial facial expression recognition technology has raised concerns both about the scientific basis of inferring emotions from facial expressions \cite{barrett2019emotional}, as well as ethical concerns surrounding the potential misuses of such technology \cite{crawford2019ai, ong2021ethical, hernandez2021guidelines}. 
The lack of systematic investigations---like this study---on the out-of-domain generalizability of deep learning models, a prerequisite for scientific validity, should be another cause for worry. Researchers all know that transfer performance on out-of-distribution examples will suffer, and by definition all deployed technology operates in this regime, where new data comes from a different distribution as the data the models were trained on. 
But, until the present extensive study, we do not know just how much performance \emph{really} suffers for facial expression recognition. 
Importantly, such technologies are already being deployed in a variety of applications, and if we cannot guarantee that our AI models can accurately recognize facial expressions even on a mix of well-controlled and curated datasets, how much confidence can we have that they will work when deployed?

Ironically, the incentives in academic publishing are even set up against these types of investigations, because such well-controlled systematic research using existing models and existing datasets are deemed ``not novel". As a field, we need to ``slow down" to conduct these extensive investigations to systematically assess the generalizability of our technologies as it is a necessary pre-requisite for ethical deployment \cite{ong2021ethical}. But if we look outside academia and consider commercially-deployed products, there are even less incentives for such systematic investigations---especially if the results suggest that more years of research are needed before revenue can start coming in. We need to pressure commercial offerings to similarly demonstrate such generalizability. One possible solution is through government regulation or auditing, for example creating a parallel model to the US National Institutes of Standards and Technology's Facial Recognition Vendor Test (FRVT) program to assess potential vendors of facial recognition technology. Another solution that could work through academic or non-profit partners would be to create audit or ``accreditation" programs with carefully curated (and secret) data that ensures the desired diversity (e.g., of demographics or emotions). Note that we think these are possible solutions that may not, by themselves, be \emph{sufficient} to deem facial expression technology ``ethical" by any means: these are just to ensure one \emph{necessary} criterion, which is that of generalizability.

As we were in the middle of conducting this research, Microsoft Azure announced in a blogpost \cite{sarah@microsoft} on 21 June 2022, that they were ceasing to offer their emotion recognition API services, citing as one of their reasons the ``inability to generalize the linkage between facial expression and emotional state across use cases, regions, and demographics". Although the service stoppage directly impacted the present study (producing in some missing values in Experiment 3), we agree with their assessment. We hope that our study will also prompt other companies to reconsider the validity of these API offerings, and to constantly evaluate and assess their offerings in light of scientific evidence.




\section{  Conclusion   }

We presented the largest-to-date systematic empirical evaluation of domain generalization in facial expression recognition using twelve datasets with single-source and multi-source training setups. Our analysis shows that single-source, single-target generalization is poor. We can improve the performance by training models on multiple source datasets, but even in our best-performing settings (train on eleven datasets and test on one dataset), we find that mean seven-class classification performance (of $\sim$65.6\%) is still significantly lower than the corresponding within-corpus accuracies ($\sim$76.4\%), with a mean performance drop of 10.8 percentage points! Commercially-available products also underperform the within-corpus accuracies by $\sim$25 percentage points. Real-world performance on novel data will almost certainly be worse, especially when we consider other sources of variability like extensions to other emotions (beyond these most commonly-studied emotions), diverse cultures and demographic groups (beyond those represented in these datasets), and so forth. We urge a great deal of caution when making or evaluating assertions about the generalizability of current and future facial expression recognition models.


\ifCLASSOPTIONcompsoc
  \section*{Acknowledgments}
\else
  \section*{Acknowledgment}
\fi

This project was supported by the National Research Foundation Singapore and DSO National Laboratories under the AI Singapore Programme (AISG Award No: AISG2-RP-2020-016), and a National University of Singapore Start-Up Grant to DCO. 
We thank Kong Yan San, Jonathon Soh, Jing Lin, and Deng Jingyuan for their help on an earlier version of this project. 
We acknowledge the use of different facial expression recognition datasets, including the IASLab\footnote{Development of the Interdisciplinary Affective Science Laboratory (IASLab) Face Set was supported by the National Institutes of Health Director’s Pioneer Award (DP1OD003312) to Lisa Feldman Barrett. More information is available online at www.affective-science.org.}, in accordance with their EULAs.

\ifCLASSOPTIONcaptionsoff
  \newpage
\fi



\bibliographystyle{IEEEtran}
\bibliography{biblio}

\begin{IEEEbiography}[{\includegraphics[width=1in,height=1.25in,clip,keepaspectratio]{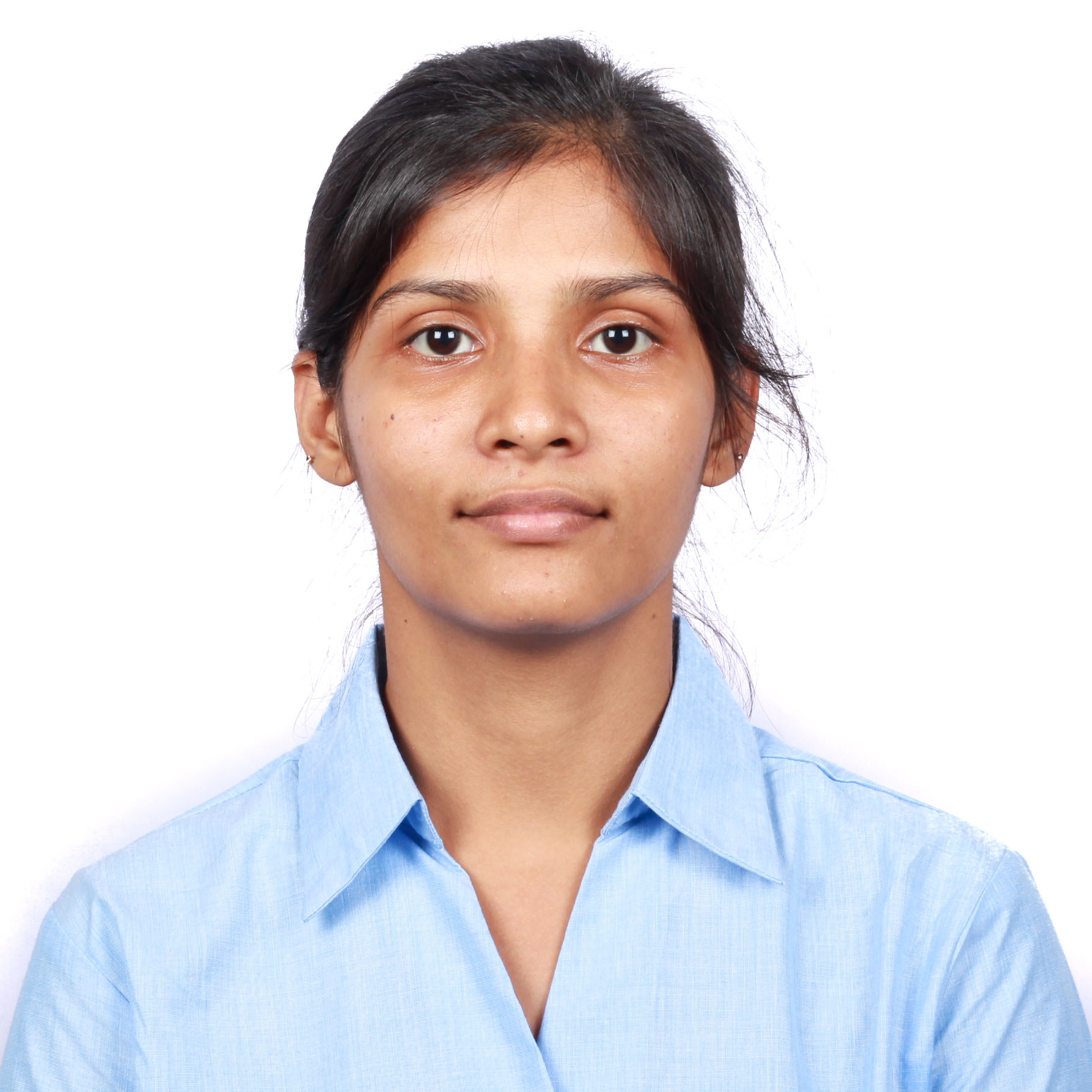}}]{Varsha Suresh} is a Ph.D. student at National University of Singapore. She received her B.Tech in Electronics and Communication Engineering from NIT Trichy in 2018. Her research interests include developing context-aware deep learning models to tackle problems like recognising fine-grained classes and improving generalisation capability of deep learning models, with application to affective computing.
\end{IEEEbiography}

\begin{IEEEbiography}[{\includegraphics[width=1in,height=125in,clip,keepaspectratio]{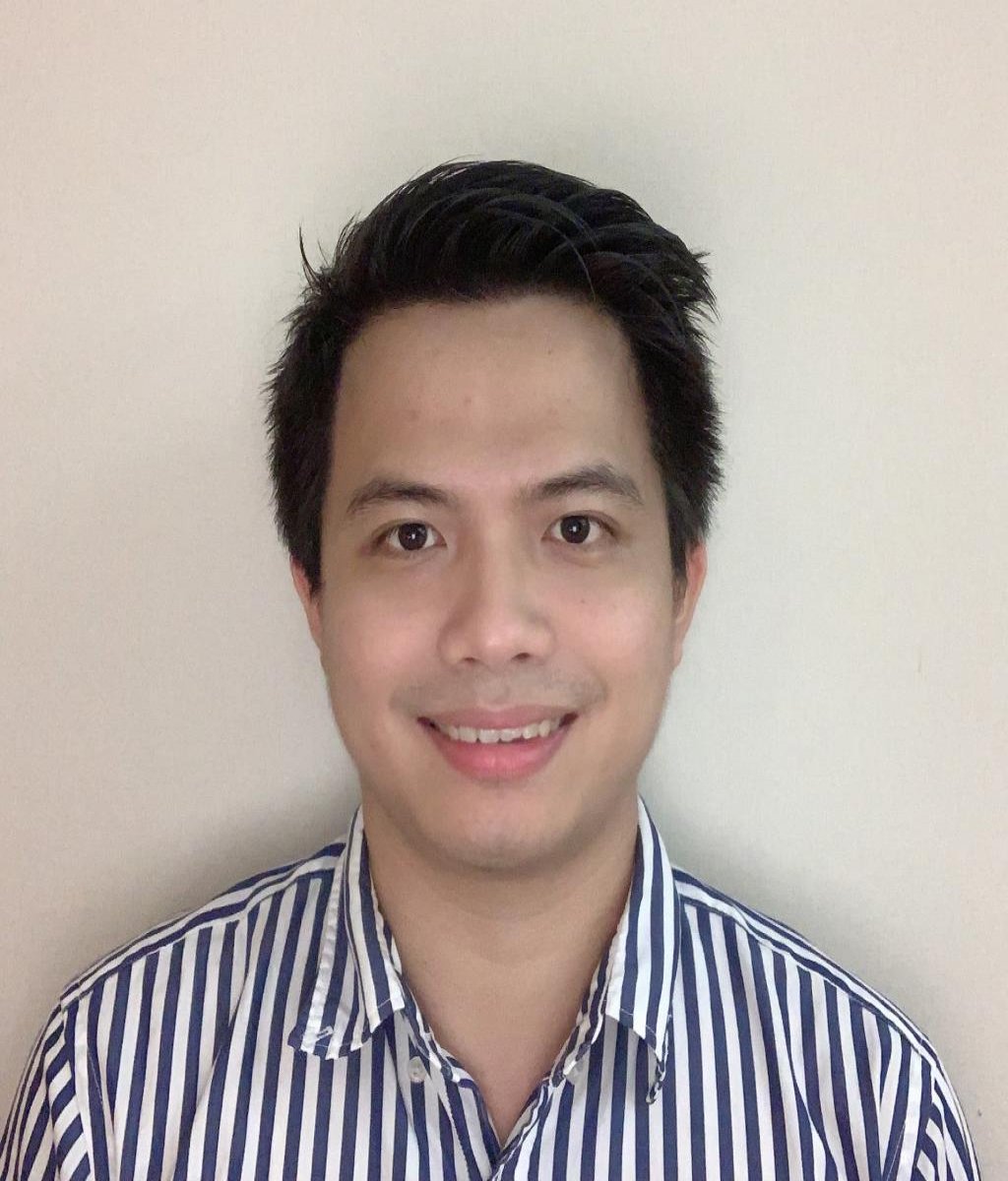}}]{Gerard Yeo} is a Ph.D student at the Institute of Data Science, National University of Singapore. He obtained his M.Soc.Sci. and B.Soc.Sci in 2018 at the National University of Singapore. Prior to his Ph.D studies, Gerard worked as a research officer under the Ministry of Social and Family Development, Singapore. His research interests include building computational models that integrate theories from Psychology and machine learning to understand emotional experiences of people, and also how people make sense of emotional situations of others.
\end{IEEEbiography}



\begin{IEEEbiography}[{\includegraphics[width=1in,height=1.25in,clip,keepaspectratio]{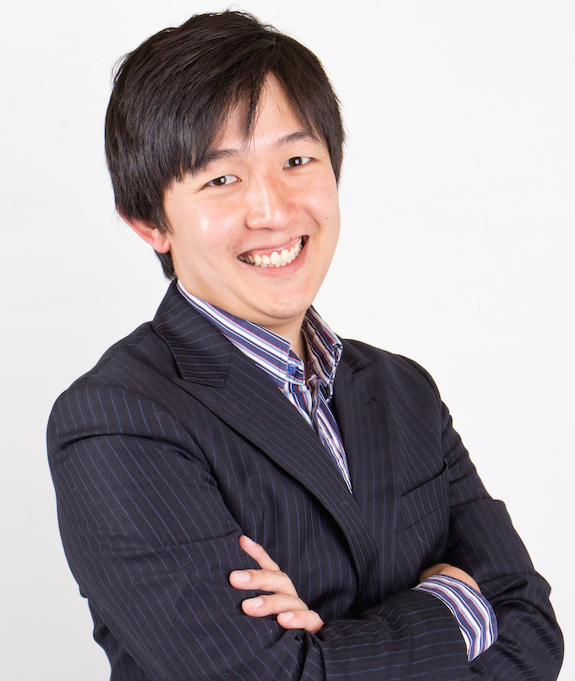}}]{Desmond C. Ong} is an assistant professor of psychology at the University of Texas at Austin. Prior to joining UT, he was a faculty member in the School of Computing at the National University of Singapore, and a research scientist at the Institute of High Performance Computing, A*STAR Singapore.
Desmond received his Ph.D. in Psychology and M.Sc. in Computer Science in 2017 from Stanford University. He graduated with a B.A. in Economics (\textit{summa cum laude}) and Physics (\textit{magna cum laude}), with minors in Cognitive Studies and Information Science from Cornell University in 2011. 
His research interests include building computational models of emotion and mental state understanding, using a mix of human behavioral experiments and modeling approaches like probabilistic modeling and machine learning. He is a member of the IEEE Computer Society.
\end{IEEEbiography}

\newpage

\appendices
\begin{table*}[tbh!]
\centering
\resizebox{\textwidth}{!}{
\begin{tabular}{l|cccccccccccc}
                        & \multicolumn{12}{c}{\textbf{Target dataset}}  \\
\textbf{Source dataset} & JAFFE              & GEMEP              & CK+                & Oulu-CASIA         & KDEF               & IASLab             & EmotioNet          & SFEW               & RAFDB              & AffWild2           & FER2013            & AffectNet           \\ 
\hline
JAFFE                   & \uline{80.9 (15.2)} & 25.4 (5.2)          & 28.6 (5.5)          & 20.0 (4.6)          & 35.5 (7.2)          & 28.5 (3.0)          & 35.1 (14.0)         & 15.9 (1.7)          & 21.5 (5.8)          & 13.4 (4.8)          & 20.0 (4.8)          & 20.7 (2.0)           \\
GEMEP                   & 14.5 (5.9)          & \uline{71.5 (6.4)}  & 14.1 (3.0)          & 15.7 (2.3)          & 25.9 (3.1)          & 20.0 (1.5)          & 24.9 (2.2)          & 20.0 (2.0)          & 21.4 (2.8)          & 15.6 (1.8)          & 19.3 (1.9)          & 18.9 (1.3)           \\
CK+                     & 29.1 (4.1)          & 13.8 (3.4)          & \uline{88.6 (4.3)}  & \textbf{54.6 (8.3)}          & 59.4 (3.8)          & 47.6 (5.2)          & 61.1 (4.0)          & 24.2 (0.4)          & 45.4 (3.4)          & 26.8 (8.1)          & 30.0 (4.0)          & 34.2 (1.1)           \\
Oulu-CASIA              & 31.8 (0.0)          & 8.5 (6.9)           & 66.8 (3.4)          & \uline{59.6 (2.7)}  & 53.7 (2.2)          & 45.0 (3.5)          & 49.6 (9.1)          & 16.2 (1.9)          & 30.7 (3.1)          & 15.9 (1.2)          & 21.5 (3.7)          & 30.4 (2.2)           \\
KDEF                    & 38.2 (10.0)         & 22.3 (1.7)          & \textbf{83.2 (2.6)} & 44.6 (5.5)          & \uline{89.2 (1.2)}  & 64.2 (2.2)          & 65.3 (5.8)          & 23.3 (4.0)          & 42.2 (3.5)          & 27.6 (6.6)          & 23.6 (5.5)          & 37.0 (1.3)           \\
IASLab                  & 35.5 (10.9)         & 16.9 (7.0)          & 63.2 (7.3)          & 44.6 (2.8)          & 73.1 (1.5)          & \uline{90.4 (1.1)}  & 69.1 (2.0)          & 26.4 (2.3)          & 40.9 (2.6)          & 28.6 (4.0)          & 23.4 (4.2)          & 38.4 (1.1)           \\
EmotioNet               & 22.7 (5.6)          & 23.1 (2.7)          & 50.5 (3.0)          & 47.5 (3.7)          & 54.3 (2.5)          & 47.9 (1.7)          & \uline{88.2 (0.8)}  & 26.0 (1.0)          & 41.9 (1.4)          & 20.6 (1.8)          & 32.3 (2.0)          & 35.3 (0.9)           \\
SFEW                    & 20.9 (4.1)          & 29.2 (6.4)          & 51.8 (7.1)          & 26.1 (1.0)          & 40.0 (5.7)          & 40.3 (5.3)          & 54.2 (3.9)          & \uline{42.2 (1.7)}  & 42.6 (1.9)          & 30.4 (4.8)          & 33.9 (1.3)          & 32.3 (1.1)           \\
RAFDB                   & 42.7 (4.1)          & 30.8 (6.1)          & 75.9 (4.4)          & 51.1 (1.0)          & 62.7 (3.1)          & 63.3 (2.6)          & 75.8 (1.2)          & 48.5 (1.4)          & \uline{83.2 (0.4)}  & 48.5 (1.6)          & 51.5 (0.4)          & \textbf{44.5 (0.6)}  \\
AffWild2                & 19.1 (5.0)          & 12.3 (5.7)          & 55.0 (6.3)          & 26.8 (2.2)          & 33.3 (3.0)          & 30.7 (4.7)          & 41.6 (3.5)          & 31.1 (2.3)          & 50.5 (2.1)          & \uline{55.0 (1.2)}  & 39.1 (1.2)          & 29.4 (1.2)           \\
FER2013                 & 28.2 (3.8)          & 32.3 (4.4)          & 50.9 (3.0)          & 41.8 (2.0)          & 51.0 (4.0)          & 47.4 (2.8)          & 66.7 (2.9)          & 42.4 (1.3)          & 60.8 (0.8)          & 48.6 (1.6)          & \uline{67.0 (0.7)}  & 41.0 (0.5)           \\
AffectNet               & \textbf{43.6 (2.5)} & \textbf{33.1 (7.0)} & 80.9 (3.8)          & 52.9 (3.9) & \textbf{80.4 (1.8)} & \textbf{78.2 (1.7)} & \textbf{82.3 (0.4)} & \textbf{51.6 (1.0)} & \textbf{77.9 (0.8)} & \textbf{58.3 (2.2)} & \textbf{57.6 (0.4)} & \uline{55.6 (1.1)}   \\ 
\hline\hline
Minimum Difference      & -37.3              & -38.4              & -5.4               & -5                 & -8.8               & -12.2              & -5.9               & 9.4                & -5.3               & 3.3                & -9.4               & -11.1               \\
Average Difference      & -51.2              & -49                & -32.2              & -20.9              & -37.4              & -43.8              & -31.3              & -12.6              & -39.9              & -24.6              & -35                & -22.7               \\
\hline
\end{tabular}
}
\vspace{0.1cm}
\caption{\textbf{Replication of Experiment 1 with a second model.} Results of single-source, single-target experiment using Inception-ResNet. Values indicate the average of seven-class classification accuracy over 5 runs, with standard deviation given in parentheses. Rows correspond to source datasets, and columns to target datasets. Diagonal entries (underlined) indicate the reference performance when model is trained and tested on the same dataset (i.e., within-corpus performance) while the off-diagonal entries are for cross-corpus to the target dataset. Best-performing values for each column, excluding the diagonals, are bolded. Models trained on GEMEP and EmotioNet as source did not see any \emph{neutral} class during training. The last two rows give the minimum difference in cross-corpus performance compared to within-corpus (i.e., bolded values $-$ underlined) and the averaged difference in cross-corpus performance compared to within-corpus (i.e., average of non-underlined values $-$ underlined).}
\label{tab:inception-resnet_single_source}
\end{table*}

\begin{table*}
\centering
\resizebox{\textwidth}{!}{
\begin{tabular}{l|cccccc||cccccc}
\multicolumn{1}{c|}{}                         & \multicolumn{12}{c}{\textbf{Target Dataset}}                                                                                                                                                                                                                                                                                                                                        \\
\multicolumn{1}{c|}{\textbf{Source Datasets}} & JAFFE                         & GEMEP                        & CK+                          & Oulu-CASIA                   & KDEF                         & IASLab                       & EmotioNet                    & SFEW                         & RAF-DB                       & Aff-Wild2                    & FER-2013                     & AffectNet                     \\ 
\hline
\textit{In-Lab Datasets ...}                  & \multicolumn{6}{c||}{Within-setting}                                                                                                                                                     & \multicolumn{6}{c}{Cross-setting}                                                                                                                                                        \\
... excluding JAFFE                           & 47.3 (5.2)                    & -                            & -                            & -                            & -                            & -                            & 73.3 (2.0)                   & 30.5 (2.6)                   & 50.8 (1.6)                   & 32.0 (3.5)                   & 35.5 (2.2)                   & 42.2 (0.8)                    \\
... excluding GEMEP                           & -                             & 20.0 (3.2)                   & -                            & -                            & -                            & -                            & 74.4 (3.4)                   & 26.6 (0.9)                   & 48.1 (1.4)                   & 29.1 (3.4)                   & 31.1 (2.8)                   & 42.6 (0.6)                    \\
... excluding CK+                             & -                             & -                            & 83.6 (4.1)                   & -                            & -                            & -                            & 73.2 (2.7)                   & 31.0 (2.7)                   & 49.1 (1.0)                   & 29.7 (3.8)                   & 33.5 (2.0)                   & 40.6 (1.1)                    \\
... excluding Oulu-CASIA                      & -                             & -                            & -                            & 54.6 (4.5)                   & -                            & -                            & 73.5 (3.3)                   & 29.3 (2.3)                   & 49.0 (1.8)                   & 33.0 (5.1)                   & 32.3 (3.6)                   & 40.6 (1.3)                    \\
... excluding KDEF                            & -                             & -                            & -                            & -                            & \textbf{81.0 (1.9)}          & -                            & 73.9 (1.9)                   & 28.9 (1.0)                   & 49.6 (1.5)                   & 31.3 (4.6)                   & 35.6 (2.0)                   & 40.4 (0.8)                    \\
... excluding IASLab                          & -                             & -                            & -                            & -                            & -                            & 66.1 (2.5)                   & 72.1 (3.7)                   & 28.3 (1.9)                   & 48.3 (1.6)                   & 24.0 (3.1)                   & 31.9 (3.5)                   & 39.5 (1.0)                    \\
                                              & \multicolumn{6}{c||}{}                                                                                                                                                                   & \multicolumn{6}{c}{}                                                                                                                                                                     \\
\textit{In-the-wild Datasets ... }            & \multicolumn{6}{c||}{Cross-setting}                                                                                                                                                      & \multicolumn{6}{c}{Within-setting}                                                                                                                                                       \\
... excluding EmotioNet                       & 47.3 (2.5)                    & \textbf{36.9 (3.4)}          & 83.2 (2.0)                   & 61.8 (4.7)                   & 75.9 (2.1)                   & \textbf{80.3 (1.8)}          & \textbf{80.2 (1.0)}          & -                            & -                            & -                            & -                            & -                             \\
... excluding SFEW                            & 47.3 (2.5)                    & 36.9 (8.0)                   & 83.2 (3.8)                   & \textbf{63.6 (3.0)}          & 75.7 (2.2)                   & 78.5 (2.3)                   & -                            & \textbf{52.0 (2.4)}          & -                            & -                            & -                            & -                             \\
... excluding RAF-DB                          & 42.7 (4.1)                    & 34.6 (2.7)                   & 84.5 (4.4)                   & 52.9 (2.4)                   & 79.2 (2.2)                   & 77.6 (2.8)                   & -                            & -                            & \textbf{77.6 (0.5)}          & -                            & -                            & -                             \\
... excluding Aff-Wild2                       & 44.5 (3.8)                    & 36.2 (5.2)                   & \textbf{84.5 (1.9)}          & 63.6 (5.0)                   & 76.7 (1.3)                   & 79.0 (1.2)                   & -                            & -                            & -                            & \textbf{59.3 (1.0)}          & -                            & -                             \\
... excluding FER-2013                        & \textbf{49.1 (2.0)}           & 36.2 (3.4)                   & 82.3 (1.9)                   & 63.2 (3.0)                   & 76.1 (1.9)                   & 76.5 (1.5)                   & -                            & -                            & -                            & -                            & \textbf{55.7 (0.6)}          & -                             \\
... excluding AffectNet                       & 42.7 (2.5)                    & 33.8 (3.2)                   & 77.7 (1.9)                   & 59.6 (4.1)                   & 61.0 (4.0)                   & 59.2 (0.6)                   & -                            & -                            & -                            & -                            & -                            & \textbf{47.9 (0.3)}           \\ 
\hline
Within-corpus performance                     & \multirow{2}{*}{80.9  (15.2)} & \multirow{2}{*}{71.5  (6.4)} & \multirow{2}{*}{88.6  (4.3)} & \multirow{2}{*}{59.6  (2.7)} & \multirow{2}{*}{89.2  (1.2)} & \multirow{2}{*}{90.4  (1.1)} & \multirow{2}{*}{88.2  (0.8)} & \multirow{2}{*}{42.2  (1.7)} & \multirow{2}{*}{83.2  (0.4)} & \multirow{2}{*}{55.0  (1.2)} & \multirow{2}{*}{67.0  (0.7)} & \multirow{2}{*}{55.6  (1.1)}  \\
from Exp. 1                                   &                               &                              &                              &                              &                              &                              &                              &                              &                              &                              &                              &                               \\
Minimum difference                            & -31.8                         & -34.6                        & -4.1                         & 4.0                          & -8.2                         & -10.1                        & -8.0                         & 9.8                          & -5.6                         & 4.3                          & -11.3                        & -7.7                          \\
\hline
\end{tabular}
}
\vspace{0.1cm}
\caption{\textbf{Replication of Experiment 2 with a second model.} Results of the Within-Setting and Cross-Setting conditions using Inception-ResNet pre-trained on CASIA-WebFace, where each row indicates a model that is trained on five datasets from the same setting (i.e., five in-lab or five in-the-wild datasets). 
Values indicate the average of seven-class classification accuracy over 5 runs, with standard deviation given in parentheses.}
\label{tab:inception-resnet_within_cross_setting}
\end{table*}

\begin{table}
\centering
\begin{tabular}{l|ccc}
& \multicolumn{3}{c}{Leave-one-out Multi-Source}  \\
Dataset             & Acc       & F1        & $\Delta$ Acc                   \\ 
\hline
JAFFE      & 48.2 (2.5) & 37.0 (1.9) & -32.7                        \\
GEMEP      & 39.2 (5.0) & 40.3 (5.9) & -32.3                       \\
CK+        & 86.4 (1.6) & 85.8 (1.2) & -2.2                           \\
Oulu-CASIA & 66.4 (5.6) & 62.0 (6.8) & 6.8                            \\
KDEF       & 83.5 (3.6) & 83.1 (3.8) & -5.7                          \\
IASLab     & 80.7 (2.5) & 79.8 (2.6) & -9.7                          \\
EmotioNet  & 81.2 (1.8) & 80.9 (2.4) & -7.0                          \\
SFEW       & 51.2 (0.7) & 47.6 (0.7) & 9.0                       \\
RAF-DB     & 77.8 (0.4) & 77.5 (0.5) & -5.4                         \\
Aff-Wild2  & 58.3 (1.0) & 57.3 (0.9) & 3.3                          \\
FER-2013   & 56.3 (0.5) & 53.0 (0.6) & -10.7                        \\
AffectNet  & 48.2 (1.3) & 46.4 (1.5) & -7.4                        \\
\hline
\end{tabular}
\vspace{0.1cm}
\caption{\textbf{Replication of Experiment 2: Leave-one-out condition using a second model.} The first column gives the classification accuracy, the second column gives the F1 score, and the third column gives the difference in accuracy compared to the within-corpus results.}
\label{tab:summary_inception_resnet}
\end{table}

\section{Repeating Analysis with Inception-ResNet model}

In the main text, we reported the results of using a ResNet50 model pre-trained on VGGFace2 and fine-tuned on the facial expression recognition task. In Tables \ref{tab:inception-resnet_single_source}, \ref{tab:inception-resnet_within_cross_setting} and \ref{tab:summary_inception_resnet}, we reproduce the results of Experiments 1 and 2 using a second model, Inception-ResNet \cite{szegedy2017inception} pre-trained with CASIA-WebFace \cite{yi2014learning}. The results qualitatively replicate the findings from the main text, with only small numerical differences.

In Table \ref{tab:summary_inception_resnet} we also give the the weighted F1-score:
\begin{equation}
    \text{weighted F1} = 2 \, \sum_{l}\frac{N_{l}}{N_\text{total}}\frac{\text{precision}_{l} \times \text{recall}_{l}}{\text{precision}_{l} + \text{recall}_{l}}
\end{equation}

\noindent where $N_{l}$ is the number of samples in class $l$ and $N_{\text{total}}$ is the total number of samples being evaluated.

\section{Adding Domain Generalisation}
\label{appendix:domain_generalisation}
In the main text, the ResNet50 model had no explicit consideration of domain generalization, in order to simulate the performance of real-life applications that do not explicitly consider generalization.

In this section, we applied \cite{zhao2020domain}'s approach of domain generalisation using entropy regularisation on ResNet50, and adapted it to facial expression recognition. Along with the standard Cross-Entropy Loss, they have an additional three losses---one which uses adversarial learning to "unlearn" the domain discriminative features and the other two focuses on Entropy Regularisation. The weighting factors of these losses were set to $\alpha_{1}$: 0.1, $\alpha_{2}$: 0.001 and $\alpha_{3}$: 0.05. To ensure a fair comparison, we use the same data-prepossessing steps and model-training steps as in the main text. We note that the single-source version of this model collapses to the ResNet50 model in the main text

The results are reported in Table \ref{tab:summary_with_domain_generalisation}. We report only results from the multi-source leave-one-out, and we compare with the results of the model without domain generalization (i.e., the model in the main text). 
We can see that domain generalization improves the cross corpus classification accuracies on some datasets but not others; the mean improvement is smaller than 0.1 percentage points (range: $\sim$ -3.3 to +3.9 \emph{pp.}).

\begin{table}
\centering
\resizebox{\columnwidth}{!}{
\begin{tabular}{lccc|ccc}
& \multicolumn{3}{c|}{Exp 2: Leave-one-out Multi-Source} & \multicolumn{3}{c}{Exp 2: Leave-one-out Multi-Source}  \\
&            \multicolumn{3}{c|}{without Domain Generalistion} & \multicolumn{3}{c}{with Domain Generalistion}        \\
Dataset             & Acc       & F1        & $\Delta$Acc~                  & Acc       & F1        & $\Delta$ Acc                   \\ 
\hline
JAFFE     & 51.8 (8.3) & 42.4 (8.2) & -39.1                        & 53.6 (5.0) & 45.3 (7.4) & -37.3                         \\
GEMEP      & 34.6 (6.1) & 38.0 (6.6) & -48.5                        & 38.5 (4.7) & 42.7 (4.3) & -44.6                         \\
CK+        & 88.6 (2.8) & 88.6 (2.7) & -2.3                          & 89.5 (1.2) & 89.7 (1.6) & -1.4                           \\
Oulu-CASIA & 64.6 (4.6) & 61.7 (5.2) & -0.8                          & 63.9 (2.0) & 59.4 (2.4) & -1.5                           \\
KDEF       & 83.9 (2.8) & 82.6 (3.8) & -5.9                         & 82.7 (1.9) & 81.7 (2.5) & -7.1                          \\
IASLab     & 80.4 (2.0) & 79.7 (2.0) & -10.8                        & 77.1 (3.0) & 76.5 (2.9) & -14.1                         \\
EmotioNet  & 82.5 (1.8) & 82.3 (1.9) & -6.6                         & 83.2 (1.0) & 82.7 (0.9) & -5.9                          \\
SFEW       & 54.7 (2.6) & 51.4 (2.7) & 9.0                          & 53.0 (0.6) & 49.7 (0.7) & 7.3                           \\
RAF-DB     & 78.6 (1.6) & 78.5 (1.4) & -6.7                         & 79.2 (0.6) & 78.9 (0.5) & -6.1                          \\
Aff-Wild2  & 60.3 (0.6) & 59.7 (0.6) & 1.8                          & 60.2 (0.7) & 59.5 (0.5) & 1.7                           \\
FER-2013   & 58.5 (0.7) & 55.7 (0.7) & -12.0                        & 59.1 (0.7) & 56.3 (1.1) & -11.4                         \\
AffectNet  & 48.6 (0.5) & 46.7 (0.5) & -7.2                         & 47.8 (0.5) & 45.9 (0.7) & -8.0                          \\
\hline
\end{tabular}
}
\vspace{0.1cm}
\caption{ResNet50 results without Domain Generalizaion (i.e., the model in the main text) on the left, and with Domain Generalisation on the right. The differences in accuracies are with the within-corpus results of the ResNet50 model. 
}
\label{tab:summary_with_domain_generalisation}
\end{table}

\end{document}